\documentclass[lettersize,journal]{IEEEtran}
\usepackage{amsmath,amsfonts}
\usepackage{algorithm} 
\usepackage{algpseudocode}
\usepackage{array}
\usepackage{textcomp}
\usepackage{stfloats}
\usepackage{subcaption}
\usepackage{url}
\usepackage{verbatim}
\usepackage{graphicx}
\usepackage{cite}
\usepackage[table,dvipsnames]{xcolor}
\usepackage{booktabs}
\usepackage{threeparttable}
\usepackage{siunitx}
\usepackage{adjustbox}
\usepackage{float}
\usepackage{amssymb}  
\usepackage{pifont}
\usepackage{xcolor}
\usepackage{tikz}
\usepackage{wrapfig}
\usepackage{makecell}
\usepackage{diagbox}
\usepackage{tabularx}
\usepackage{multirow}
\usepackage{tcolorbox} 
\usepackage[hidelinks]{hyperref}
\usepackage[textsize=tiny]{todonotes}
\usepackage{listings}
\begin{document}
\title{RAG-FGO: Dynamic Optimization of Query-Rewriting Agents for Generative RAG}

\author{Qigan Sun, Xudong Wang, Chaoning Zhang~\IEEEmembership{Senior Member, IEEE,}, Chenghao Li, Jiaquan Zhang, Jiehui Xie, Sungyoung Lee,~\IEEEmembership{Member, IEEE,} Jie Zou, Ning Xie, Peng Wang, Yang Yang,~\IEEEmembership{Senior Member, IEEE,} Heng Tao 
Shen,~\IEEEmembership{Fellow, IEEE}

\thanks{Qigan Sun, Xudong Wang, and Sungyoung Lee are with the School of Computing, Kyung Hee University, Yongin-si, South Korea (e-mail: sunqigan0206@gmail.com; wl200203@khu.ac.kr; sylee@oslab.khu.ac.kr).}
\thanks{Chaoning Zhang, Chenghao Li, Jiaquan Zhang, Jiehui Xie, Ning Xie, Jie Zou, Peng Wang, and Yang Yang are with the School of Information and Software Engineering, University of Electronic Science and Technology of China, Chengdu, China (e-mail: chaoningzhang1990@gmail.com; lch17692405449@gmail.com; 2024091202001@std.uestc.edu.cn; jiehuixie@std.uestc.edu.cn; jie.zou@uestc.edu.cn;  seanxiening@gmail.com; p.wang6@hotmail.com; yang.yang@uestc.edu.cn).}

\thanks{Heng Tao Shen is with the School of Computer Science and Technology, Tongji University, Shanghai, China (e-mail: shenhengtao@hotmail.com).}

% \thanks{* Corresponding Author}
}


\maketitle

\begin{abstract}
In retrieval-augmented generation (RAG), query rewriting serves as a core interface that aligns user intent with external knowledge bases, and its effectiveness directly determines overall system performance. However, generative retrieval faces three critical challenges in real-world deployment: noisy retrieval feedback, the difficulty of jointly optimizing discrete prompt templates and continuous decoding-control parameters, and the limited generalization of static rewriting strategies under semantically complex scenarios. To address these issues, we propose RAG-FGO, a dynamic optimization framework for generative retrieval strategies. RAG-FGO formulates query rewriting as a non-differentiable black-box optimization problem in a mixed discrete–continuous space, and represents prompt templates and decoding parameters as adaptively optimizable query-rewriting agents. The framework performs global search to enable semantic jumps and structural reorganization that escape local optima, while applying local perturbations for fine-grained alignment and precise refinement. In addition, RAG-FGO introduces an elite-preserving memory pool that accumulates high-quality strategies across iterations, thereby mitigating the impact of stochasticity and noisy feedback on optimization stability. Experimental results on a diverse suite of benchmarks demonstrate the effectiveness and robustness of the proposed method, supporting its applicability to complex semantic retrieval tasks.
\end{abstract}

\begin{IEEEkeywords}
Retrieval-Augmented Generation, Query Rewriting, Agent-based Optimization
\end{IEEEkeywords}

\section{Introduction}
Retrieval-Augmented Generation (RAG) integrates external knowledge sources into the generation process by retrieving relevant evidence from a knowledge corpus and injecting it into the model context, thereby explicitly coupling language models with external memory. As a result, RAG has become a core paradigm for knowledge-intensive tasks such as open-domain question answering, fact verification, and domain-specific text generation~\cite{lewis2020retrieval,huang2024survey,pan2024unifying,hong2025next}. Compared to generation methods that rely solely on parametric memory, RAG dynamically incorporates traceable evidence at inference time, providing factual grounding for generated outputs. This design reduces hallucination caused by missing or outdated knowledge and improves the verifiability and reliability of model predictions~\cite{petroni2019language,shuster2021retrieval}. In most traditional RAG systems, however, the user query is assumed given, and the retrieval module passively matches documents based on the original query~\cite{karpukhin2020dense,pan2024unifying}.
\begin{figure}[t]
    \centering
    \captionsetup{skip=2pt}
    \includegraphics[width=0.47\textwidth]{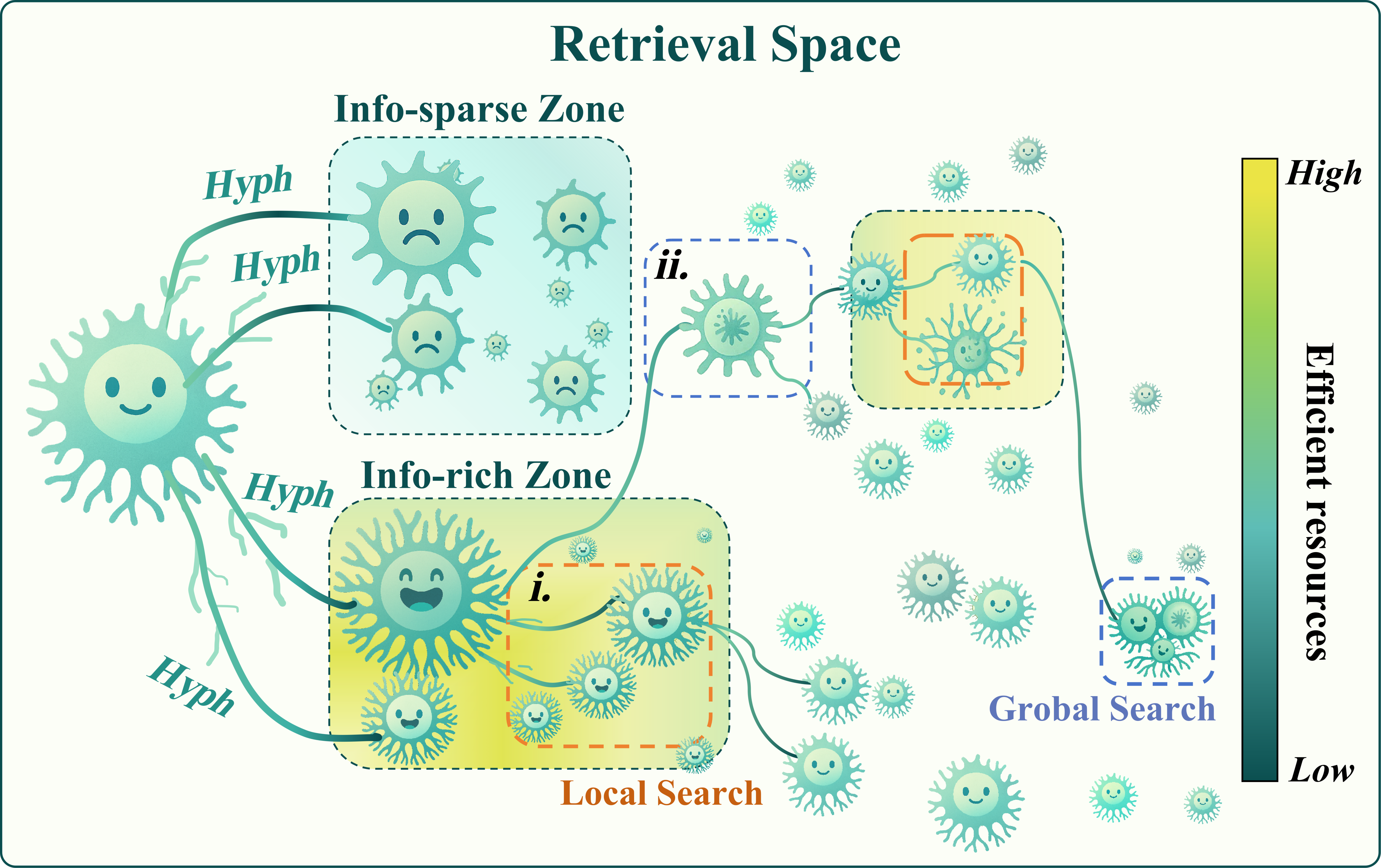}
    \caption{Conceptual illustration of the FGO-inspired search dynamics in RAG-FGO. Hyphae grow from low-efficiency (info-sparse) zones toward high-efficiency (info-rich) zones in the retrieval space. (i) Local search enables fine-grained exploitation within promising regions; (ii) global search facilitates large semantic jumps to escape local optima and discover new resource-rich areas.}
    \label{fig:concept11}
\end{figure}
Under this setting, retrieval effectiveness heavily depends on the surface form of the original query. Semantic mismatches between query expressions and knowledge corpus representations often directly limit the accessible evidence space~\cite{gao2023retrieval,gao2023precise}. To mitigate this issue, generative retrieval enhances RAG’s adaptability in complex semantic scenarios. Generative retrieval leverages large language models (LLMs) to actively rewrite and structure user intents, producing retrieval-friendly queries that better align with the knowledge corpus. This approach facilitates evidence retrieval under semantic diversity, ambiguous expressions, or cross-domain transfer~\cite{wang2023query2doc,dai2022promptagator}. More importantly, the generate-then-retrieve paradigm introduces a controllable semantic transformation space into the retrieval process, enabling the system to flexibly align with the organization and inductive biases of the knowledge corpus without relying solely on the retriever’s representation capacity or index structure~\cite{sharma2025retrieval}. However, this flexibility also introduces new challenges. The behavior of generative retrieval depends not only on the retriever but also on the generation configuration, which makes strategy optimization non-trivial. Moreover, retrieval quality is often assessed through indirect or proxy feedback signals that are noisy and may drift across queries and domains, rendering strategy selection and optimization susceptible to noise and non-stationary distributions~\cite{yang2023large,liu2024lost}. As a result, static rewriting strategies frequently fail to generalize under semantically complex scenarios.

Motivated by these observations, we adopt a different modeling perspective: \textit{Instead of treating generative retrieval as a one-shot query rewriting process, we reformulate it as an optimization problem over a retrieval strategy space.} Specifically, we represent a retrieval strategy as query-rewriting agents jointly parameterized by discrete prompt templates and continuous decoding parameters. The prompt template defines the structural form of query rewriting, while decoding parameters (e.g., temperature, top-$p$, and repetition penalty) control the generation distribution and output characteristics. Given a user query, our query-rewriting agents generate rewritten queries for retrieval. Their performance is evaluated based on the semantic similarity between the retrieved results and the target information need. Since strategy performance can only be assessed through randomly generated queries and their corresponding retrieval outcomes, and this evaluation involves non-differentiable retrieval operators and sampling randomness, the optimization problem exhibits a mixed discrete--continuous, noisy, black-box nature that is difficult to solve using gradient-based methods~\cite{zhang2025collaborative,cheng2024black}. Consequently, strategies based on static or purely local adjustments often fail to sufficiently explore the strategy space and struggle to achieve stable improvements under noisy feedback.

To address these challenges, we propose RAG-FGO, a dynamic optimization framework for query-rewriting agents. Inspired by the Fungal Growth Optimizer (FGO)~\cite{abdel2025fungal}, RAG-FGO combines global exploration with local refinement to enable effective search in complex and non-stationary strategy spaces. Figure~\ref{fig:concept11} illustrates this exploration dynamic in the retrieval strategy space. Global operations facilitate semantic jumps and structural reorganization across distant regions of the strategy space, while local operations perform fine-grained perturbations and convergent optimization over prompt semantics and decoding parameters. Furthermore, we introduce an elite-preserving memory pool that accumulates high-quality query-rewriting agents across iterations. This mechanism mitigates strategy degradation caused by generation stochasticity and noisy retrieval feedback, thereby significantly improving optimization stability. We conduct systematic evaluations on multiple knowledge-intensive question answering and reasoning benchmarks. Experimental results demonstrate that, under identical model size and retrieval budget constraints, RAG-FGO consistently improves retrieval quality and downstream task performance, and exhibits stronger robustness and transferability across semantically complex and cross-task scenarios. Our main contributions are summarized as follows:
\begin{itemize}
    \item A novel optimization perspective for generative retrieval that unifies prompt templates and decoding parameters into a single mixed discrete--continuous query-rewriting agent.
    \item The RAG-FGO dynamic optimization framework, which facilitates stable and efficient exploration of the strategy space through integrated local and global search, and an elite memory mechanism.
    \item Extensive empirical evaluations across multiple question answering and reasoning benchmarks, demonstrating the proposed method's effectiveness and generalization capability in semantically complex scenarios.
\end{itemize}

\section{Related Work}
\subsection{RAG in Large Language Models}
RAG~\cite{lewis2020retrieval} has become a widely adopted solution to the knowledge limitations of LLMs by explicitly coupling generation with external evidence retrieval. This enables more grounded and verifiable outputs in knowledge-intensive tasks such as open-domain question answering and factual verification. Prior work has advanced RAG along two main directions: enhancing retrieval quality through dense retrievers~\cite{karpukhin2020dense,jin2024large}, late-interaction models like ColBERT~\cite{khattab2020colbert}, and scalable infrastructures; and improving evidence fusion via encoder--decoder designs and retrieval-aware generation~\cite{izacard2020leveraging,mialon2023augmented}. Recent efforts further explore multi-stage pipelines~\cite{mao2021rider}, reinforcement-learning-based adaptive retrieval~\cite{jiang2023active,shi2023replug}, and long-context reasoning for larger evidence windows~\cite{bai2023longbench}.

More recently, generative retrieval has emerged as a key extension, where LLMs actively rewrite or generate queries to better align user intent with knowledge-base representations~\cite{singh2025agentic}. Representative methods include hypothetical document generation, such as HyDE~\cite{gao2023precise}, as well as LLM-based query rewriting and expansion~\cite{ma2023query}. These approaches introduce a controllable semantic transformation space and demonstrate stronger end-to-end performance on ambiguous or compositional queries.

However, existing generative retrieval methods typically rely on static prompt templates and one-shot generation, without modeling query formulation as an optimizable process. Fixed configurations often result in redundant or mismatched retrieval~\cite{gao2023retrieval}, limiting adaptation to semantic variability across tasks, query types, or domains. In contrast to prior work that treats query generation as a fixed preprocessing step, our approach reframes it as a learnable retrieval-agent search problem (jointly optimizing discrete prompt structures and continuous decoding controls under downstream feedback) to enable dynamic generative retrieval.

% \subsection{Black-box Optimization for LLM-based Systems}
\subsection{Heuristic Search Algorithms}
Heuristic and bio-inspired search algorithms are classical yet effective tools for tackling high-dimensional and non-convex optimization problems, especially when the objective function is expensive, discontinuous, or lacks analytical gradients~\cite{talbi2009metaheuristics,blum2003metaheuristics,costa2017efficient}. With the increasing reliance on black-box LLMs, similar optimization challenges naturally arise in LLM-related settings, where gradients are unavailable, unreliable, or practically inaccessible due to discrete prompt structures and stochastic generation. Consequently, gradient-free search has become a practical paradigm for improving LLM behaviors under indirect supervision signals.

Existing studies have explored a variety of heuristic strategies for prompt-related optimization. A representative line of work focuses on prompt evolution, leveraging genetic algorithms, evolutionary strategies, or population-based search to refine prompt candidates through iterative mutation and selection~\cite{fernando2023promptbreeder,li2024promptist,yang2023large}. Another line investigates gradient-free prompt discovery or selection mechanisms, including discrete trigger search and black-box prompt construction~\cite{shin2020autoprompt,prasad2023grips}. Reinforcement-learning-based approaches further model prompting as sequential decision making, where prompts or actions are optimized according to task rewards~\cite{deng2022rlprompt}. Beyond prompt structures, heuristic optimizers have also been applied to generation-time control, where swarm-intelligence methods such as PSO or GA tune decoding parameters or steer generation behavior without backpropagation~\cite{10611913,qolomany2017parameters}.

Among these families, bio-inspired methods are particularly attractive because they tend to exhibit a favorable balance between broad exploration and fine-grained local refinement, making them robust to multi-modality and noisy feedback~\cite{yang2020nature}. The FGO~\cite{abdel2025fungal}, for instance, mimics hyphal expansion and branching to coordinate local and global search, enabling efficient navigation of complex landscapes. This search characteristic is well aligned with retrieval-agent optimization in RAG, where effective strategies often require both non-trivial semantic shifts (to escape suboptimal query formulations) and controlled refinements (to preserve intent and stability). Motivated by this alignment, we adapt FGO to the RAG setting and develop a dynamic retrieval-agent optimization framework that iteratively improves query reformulation strategies via structured exploration--exploitation in the strategy space.

\section{Method}
In this section, we present the proposed RAG-FGO framework in detail. We begin by formalizing the problem of generative retrieval optimization and defining the query-rewriting agents that jointly model prompt templates and decoding parameters (Section~\ref{sec:problem_formulation}). 
We then provide an overview of the overall framework and its optimization pipeline. (Section~\ref{sec:framework_overview}). Next, we describe the agent initialization strategy and the scoring-and-selection mechanism used to evaluate and retain high-quality query-rewriting agents (Section~\ref{sec:agent_init} and Section~\ref{sec:scoring_selection}). 
We further elaborate on the exploration–exploitation process, including local and global search operations for agent evolution (Section~\ref{sec:exploration_exploitation}). 
Finally, we introduce the elite-preserving memory pool mechanism, describe agent accumulation and termination conditions, and explain how the optimized query-rewriting agents are selected and deployed for inference (Sections~\ref{sec:memory_pool} and \ref{sec:termination-application}).

\subsubsection{Problem Formulation} 
\label{sec:problem_formulation}
In the context of RAG, the core challenge lies in bridging the semantic gap between user intent and external knowledge. 
We define a dataset $\mathcal{D} = \{(q_i, c_i^*)\}_{i=1}^N$, where $q_i$ is the user query and $c_i^*$ is the ground-truth supporting context. The knowledge base is denoted as $\mathcal{K}$. We formulate it as the optimization of query-rewriting agents $A = (C, \theta)$. Here, $C$ represents the discrete prompt template, and $\theta \in \mathbb{R}^d$ denotes the continuous decoding hyperparameters. Given an input $q$, the agent generates a rewritten query $\hat{q} = \mathcal{L}(q; A)$ via the LLM $\mathcal{L}$, which is subsequently used to retrieve context $\hat{c} = \mathcal{R}(\hat{q}, \mathcal{K})$ via the retriever $\mathcal{R}$. 

Our ultimate goal is to identify the optimal agent configuration $A^* = (C^*, \theta^*)$ that maximizes the expected semantic similarity between the retrieved evidence and the ground truth:
\begin{equation}
A^* = \arg\max_{A \in \mathcal{S}} \mathbb{E}_{(q_i, c_i^*) \sim \mathcal{D}} \left[ \text{Sim}\left( \mathcal{R}\left( \mathcal{L}(q_i; A), \mathcal{K} \right), c_i^* \right) \right],
\end{equation}
where $\mathcal{S}$ is the joint solution space of prompt templates and decoding parameters, and $\text{Sim}(\cdot, \cdot)$ is the cosine similarity metric.
\begin{figure*}[h]
    \centering
    \includegraphics[width=1\linewidth]{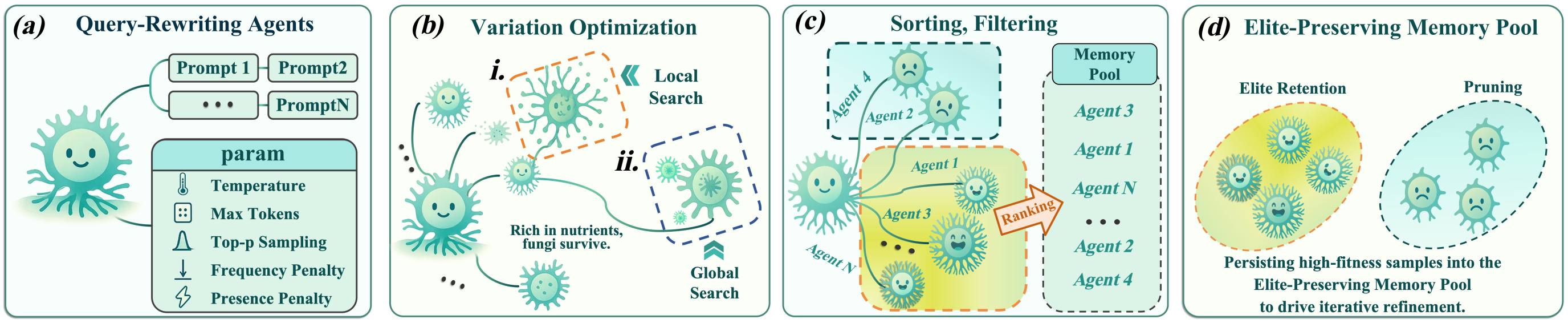}
    \caption{Overview of the main components of RAG-FGO. (a) A query-rewriting agent consists of a prompt template and continuous decoding parameters. (b) Variation optimization evolves agents via local search (synonym-level or minor phrasing edits) and global search (LLM-driven prompt rewriting). (c) Candidate agents are scored, ranked, and filtered to select elites. (d) The elite-preserving memory pool retains elite agents and prunes inferior ones across iterations for subsequent refinement.}
    \label{fig:enter-label2}
\end{figure*}

\subsubsection{Overview of RAG-FGO Framework} 
\label{sec:framework_overview}
Figure~\ref{fig:enter-label2} illustrates the key definitions and operational concepts of the proposed RAG-FGO framework. The framework is designed for the dynamic optimization of query-rewriting agents, each comprising a discrete prompt template and a set of continuous decoding parameters (e.g., temperature, top-$p$, and maximum generation length). Starting from a natural language query, the template initialization module generates an initial population of diverse agents. Each agent is used to generate a rewritten query, which is then utilized to retrieve contextual information from an external knowledge base. These retrieved results are compared against reference contexts to evaluate the performance of each agent based on semantic similarity. Following this evaluation, the query-rewriting agents are ranked, and the top-performing individuals are selected as the basis for subsequent iterative refinement.

To navigate the complex retrieval strategy space, the framework employs local search and global search in parallel, each serving a distinct optimization objective. Global search utilizes LLMs to restructure prompts entirely, facilitating large-scale semantic shifts that allow the system to escape local optima and discover new, high-efficiency zones. Conversely, local search focuses on fine-grained exploitation within promising regions by introducing subtle variations—such as synonym replacement or tone adjustment—to achieve precise alignment. The high-quality query-rewriting agents identified through these dual operations are then integrated into an elite-preserving memory pool.

\begin{figure*}[t]
    \centering
    \includegraphics[width=1\linewidth]{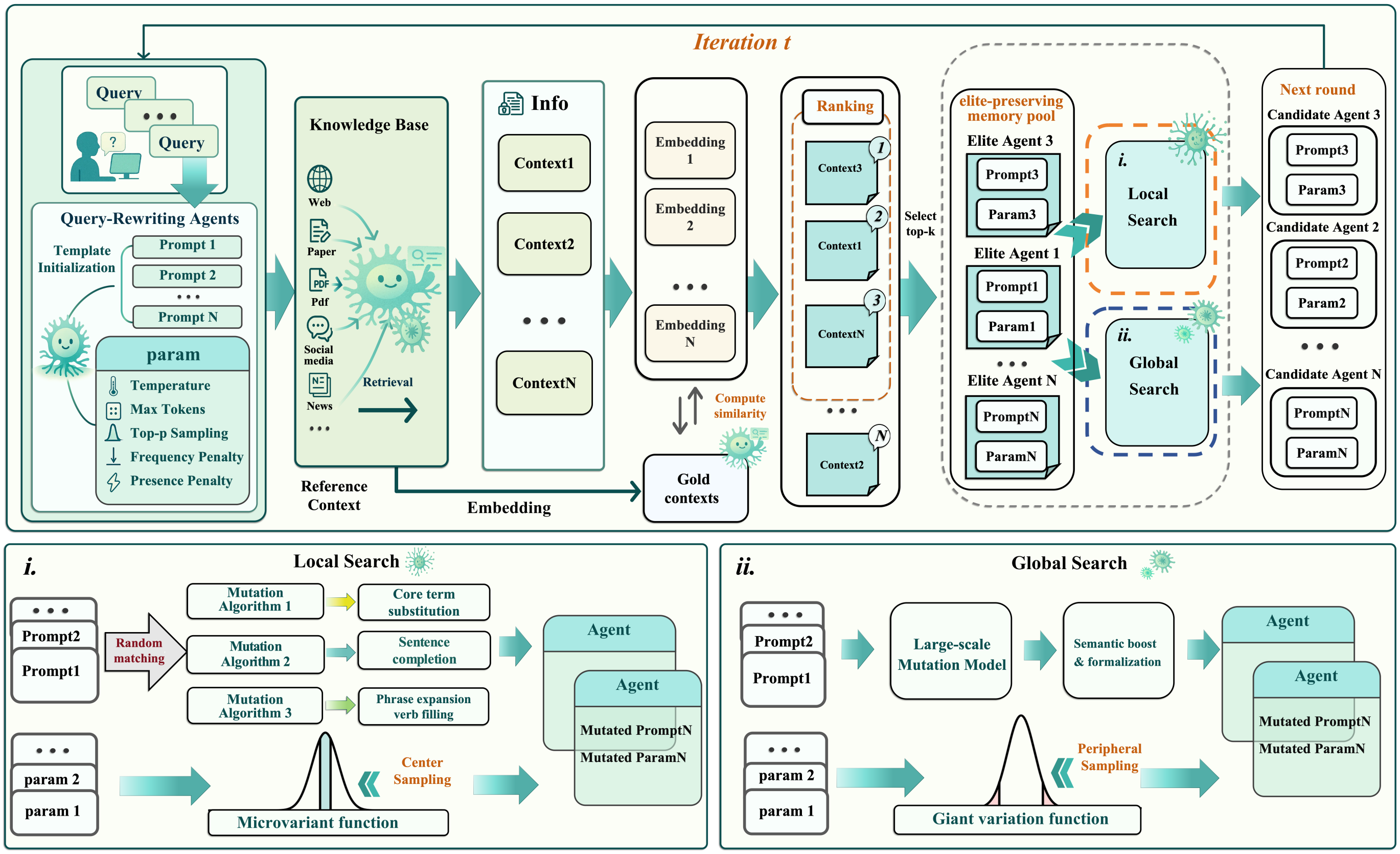}
    \caption{Training process for optimizing query-rewriting agents driven by RAG-FGO. Query-rewriting agents, consisting of a prompt template and continuous decoding parameters, iteratively rewrite queries for retrieval. Retrieved contexts are ranked by semantic similarity to reference evidence, and the corresponding elite agents are stored in an elite-preserving memory pool. The memory-retained agents are then evolved through local search (synonym-level or minor phrasing edits) and global search (LLM-driven prompt rewriting) for the next iteration.}
    \label{fig:main111}
\end{figure*}

This elite-preserving memory pool serves to mitigate the impact of retrieval noise and stochastic fluctuations by accumulating historically effective strategies over successive iterations. As the optimization progresses, the pool is continuously updated to maintain a diverse yet high-performing population of agents, providing a robust foundation for the next round of search. This cycle of exploration and memory-enhanced refinement continues until the system converges on a stable, optimal configuration. Ultimately, the agent demonstrating the highest retrieval performance is selected from the pool as the final output, ensuring significant improvements in semantic similarity and factual grounding. The detailed training workflow of RAG-FGO is illustrated in Figure~\ref{fig:main111}.

\subsubsection{Agent initialization} 
\label{sec:agent_init}
RAG-FGO first initializes a set of $N$ agents $\mathcal{A} = \{A_1, A_2, \dots, A_N\}$, where each agent $A_j = (C_j, \theta_j)$ consists of a prompt $C_j$ and parameters $\theta_j \in \mathbb{R}^d$. These agents encode diverse query augmentation strategies, covering multiple intent expressions and prompting styles to meet heterogeneous retrieval needs. Initial prompts are sampled from a predefined prompt space, which may include templates such as: \textit{``As a linguistic diversity expert, you excel at subtly rewriting sentences while preserving their core intent and information.''} In parallel, each agent is assigned a decoding parameter vector $\theta_j$ that controls generation behavior, for example: temperature = 0.7, top-$p$ = 0.8, frequency penalty = 2.0, presence penalty = 2.0, max tokens = 1024. Formally, prompts are drawn from a structured template pool, while parameters $\theta_j$ are initialized from a normal distribution to ensure diversity across the agent population:  
\begin{equation}
\theta_j \sim \mathcal{N}(\mu_0, \sigma_0^2)
\end{equation}
where $\mu_0$ and $\sigma_0$ are predefined mean and variance, respectively. This initialization ensures diversity at the starting point of optimization.

\subsubsection{Scoring and Selection} 
\label{sec:scoring_selection}
During training, we first employ the agents selected in the current iteration to generate candidate context passages, which are then compared against the corresponding reference context $c_i^*$ for semantic similarity. Specifically, the semantic similarity between a candidate context $c_{i,j}$ and its reference $c_i^*$ is then computed by measuring the cosine similarity between their corresponding embeddings, reflecting how closely the candidate aligns with the reference in terms of meaning:  
\begin{equation}
s_{i,j} = \text{Sim}(c_{i,j}, c_i^*) = \frac{c_{i,j} \cdot c_i^*}{\|c_{i,j}\| \, \|c_i^*\|} \, 
\end{equation}
Here, $c_{i,j}$ and $c_i^*$ denote dense vector representations obtained from a pretrained embedding model, which projects semantic information into a shared vector space. The resulting score $s_{i,j}$ provides a normalized measure of semantic similarity, independent of vector scale. Based on these scores, all agents are ranked within the current iteration, and the top-$k$ agents are selected according to a maximum selection rule:
\begin{equation}
\mathcal{A}_{\text{top-}k} = \operatorname{Top\mbox{-}k}\bigl(\{ (A_j, s_{i,j}) \mid A_j \in \mathcal{A} \}\bigr)
\end{equation}
This mechanism ensures that, in each iteration, only the query-rewriting agents producing candidate outputs most aligned with the reference contexts are retained. The selected agents are subsequently stored in the elite-preserving memory pool $\mathcal{M}$, which acts as a global repository of high-quality query-rewriting agents.
% By retaining the best-performing agents across iterations, the memory pool facilitates cumulative optimization, mitigates catastrophic forgetting, and provides a stable foundation for subsequent variation and diffusion processes.

\subsubsection{Exploration and Exploitation}
\label{sec:exploration_exploitation}
To maintain population diversity and prevent entrapment in local optima, RAG-FGO employs a dual-search mechanism for each selected agent $A_j = (C_j, \theta_j)$. Each agent acts as a seed to generate $mc$ (mutation count) variation candidates through iterative mutation, effectively expanding the search frontier within the strategy space. The selection between local search and global search for these $mc$ candidates is governed by a dynamic probability $p_t$. At the beginning of each iteration \(t\), a dynamic local search probability \(p_t\) is scheduled between predefined minimum and maximum values \(p_{\min}\) and \(p_{\max}\):  
\begin{equation}
p_t = p_{\min} + \frac{t-1}{T_{\max}-1}(p_{\max}-p_{\min}) \, 
\end{equation}
This scheduling strategy ensures that early iterations emphasize global search (\(p_t\) small), while later iterations gradually shift toward local refinement (\(p_t\) large).  
For each mutation, the algorithm samples from \(\mathrm{Bernoulli}(p_t)\) to decide whether to perform local search or global search. Local search introduces slight semantic modifications to the prompt and applies small perturbations to the parameters:
\begin{equation}
\begin{split}
C_j' &= m_{\text{local}}(C_j, \epsilon_c^{(l)}), \quad
\theta_j' = \theta_j + \epsilon_\theta^{(l)}, \\
\epsilon_\theta^{(l)} &\sim \text{TruncNorm}(0, \sigma_l^2, [-k_l\sigma_l, k_l\sigma_l]),
\end{split}
\end{equation}
where $m_{\text{local}}(\cdot)$ is the local modification function that performs fine-grained semantic adjustments (e.g., synonym replacement) while preserving the core intent of the prompt, and both $\epsilon_c^{(l)}$ and $\epsilon_\theta^{(l)}$ represent small perturbation magnitudes. In contrast, global search introduces more substantial modifications to both the prompt and the parameters. It excludes perturbations near zero by sampling only values whose absolute magnitude exceeds a threshold \(\delta\):
\begin{equation}
\begin{split}
C_j' = m_{\text{global}}(C_j, \epsilon_c^{(g)}), \quad
\theta_j' = \theta_j + \epsilon_\theta^{(g)}, \\
\epsilon_\theta^{(g)} \sim \text{TruncNorm}(0, \sigma_g^2, [-k_g\sigma_g, -\delta] \cup [\delta, k_g\sigma_g]), 
\end{split}
\end{equation}
where $m_{\text{global}}(\cdot)$ denotes the global restructuring function for prompt templates, which leverages a pre-trained LLM to reconstruct the prompt structure for large-scale semantic shifts without altering the retrieval task objective, while $\epsilon_c^{(g)}$ represents the perturbation intensity for global prompt variation. 
To ensure exploration, we set $\delta > 0$ to exclude trivial parameter changes and define $\sigma_g > \sigma_l$ to facilitate larger parameter shifts compared to local search. The constant $k_g$ constrains the global perturbation range to prevent excessive hyperparameter deviations, while $k_l$ serves as a local scaling coefficient that restricts the search to a narrow neighborhood of the current configuration, thereby ensuring the stability of the optimized decoding configurations throughout the refinement process.

\subsubsection{elite-preserving Memory Pool} 
\label{sec:memory_pool}
The memory pool $\mathcal{M}$ serves as a dynamic repository for retaining high-quality query-rewriting agents, ensuring that effective strategies are preserved throughout the optimization process. Formally, it maintains a population of the top-$M$ elite agents:
\begin{equation}
\mathcal{M}_t = \{A_1, A_2, \dots, A_M\}
\end{equation}
Given the non-stationary nature of query complexity across iterations, comparing candidates based on historical scores may introduce bias. To address this, RAG-FGO employs an Elite-Preserving Re-evaluation Mechanism. In each iteration $t$ with a new query batch $q_t$, the framework re-evaluates the previous elite agents $\mathcal{M}_{t-1}$ alongside the set of newly generated agents $\mathcal{A}_t$. This procedure establishes a unified performance baseline on the current task instance, enabling a direct comparison between existing and new strategies. The memory pool is subsequently updated by retaining the top-$M$ performing agents from the combined population:
\begin{equation}
\mathcal{M}_t = \operatorname{Top-M}\left( \mathcal{M}_{t-1} \cup \mathcal{A}_t \Big| q_t \right)
\end{equation}
This mechanism ensures that the retained agents represent the most effective strategies for the current retrieval context, thereby mitigating the impact of query difficulty fluctuations on the optimization trajectory.
% \subsubsection{Memory Pool} 
% \label{sec:memory_pool}
% The memory pool $\mathcal{M}$ serves as a global repository for retaining high-quality agents. Formally, it maintains the set of top-$M$ agents:
% \begin{equation}
% \mathcal{M} = \{A_1, A_2, \dots, A_M\}
% \end{equation}
% In each iteration $t$ involving a new query $q_i$, we introduce an anchor-based dynamic validation mechanism to balance evaluation fairness with computational efficiency. 
% Instead of re-evaluating the entire memory pool, we only re-evaluate the current top-ranked agent (the anchor) from $\mathcal{M}$ on $q_i$ to establish a dynamic performance baseline. 
% The newly evolved agents in the current population are then compared against this baseline. 
% If a new candidate outperforms the re-evaluated anchor, it is inserted into $\mathcal{M}$, replacing the least effective agent. 
% This design enables the long-term accumulation of diverse, high-quality prompts while ensuring that updates are driven by genuine performance improvements rather than query difficulty fluctuations.

\subsubsection{Termination and Application} 
\label{sec:termination-application}
The optimization process continues iteratively until the discovered strategy demonstrates robustness or the computational budget is exhausted. Let $A_{\text{best}}^{(t)}$ denote the highest-ranked agent in the updated elite-preserving memory pool $\mathcal{M}_t$ at iteration $t$. We define a stability counter $\tau$, which increments if the top-performing agent remains identical to that of the previous iteration (i.e., $A_{\text{best}}^{(t)} = A_{\text{best}}^{(t-1)}$) and resets otherwise. To balance optimization quality with resource efficiency, the training terminates when either of the following conditions is met:
\begin{itemize}
    \item Stabilization: The stability counter $\tau$ reaches a threshold $T_{\text{stable}}$ (indicating a robust optimum has been found).
    \item Budget Limit: The current iteration $t$ reaches a maximum limit $T_{\text{max}}$ (preventing excessive computational cost).
\end{itemize}
Upon termination, the best-performing agent from the final elite-preserving memory pool is selected as the output:
\begin{equation}
A^* = A_{\text{best}}^{(\text{final})} \in \mathcal{M}_{\text{final}}
\end{equation}
The selected agent $A^* = (C^*, \theta^*)$ contains both the optimized prompt $C^*$ and the optimized parameters $\theta^*$. This agent is then deployed in the testing phase, applying its learned rewriting strategy to unseen queries. This ensures that the system utilizes the most robust retrieval policy discovered and validated throughout the dynamic optimization process. 

% The pseudocode of the overall framework is provided in \textcolor{red}{Appendix~\ref{algo:rfgo}.}

\section{Experiments}

\subsection{Experimental Setup}
\label{Experimental_Setup}
\subsubsection{Datasets} We evaluate our methods on two groups of benchmarks. 
\textbf{QA benchmarks:} 
LongBench~\cite{bai2023longbench} targets long-context understanding with inputs up to 100K tokens. 
HotpotQA~\cite{yang2018hotpotqa} is a multi-hop QA dataset over multiple supporting documents. 
2WikiMultihopQA~\cite{ho2020constructing} focuses on compositional multi-hop questions in Wikipedia. 
MuSiQue~\cite{trivedi2022musique} decomposes multi-hop reasoning into single-hop questions. 
Natural Questions (NQ)~\cite{kwiatkowski2019natural} contains real queries with Wikipedia passages. 
SQuAD~\cite{rajpurkar2016squad} is a reading comprehension dataset with span-based answers. 
StrategyQA~\cite{geva2021did} requires implicit reasoning steps to answer strategy-based questions. AmbigQA~\cite{min2020ambigqa} targets ambiguous queries that involve multiple potential interpretations. PopQA~\cite{mallen2023not} focuses on long-tail factual knowledge with low-popularity entities. \textbf{Reasoning benchmarks:} 
MATH500~\cite{hendrycks2021measuring} is a subset of the MATH dataset containing 500 challenging competition-level problems. 
MMLU-pro~\cite{wang2024mmlu} extends the standard MMLU benchmark with more professional and domain-specific tasks. GPQA~\cite{rein2024gpqa} is a challenging benchmark that evaluates graduate-level problem-solving in science and engineering domains. DROP~\cite{dua2019drop} is a reading comprehension dataset requiring discrete numerical reasoning, such as addition, counting, subtraction, and comparison. MuSR~\cite{sprague2023musr} is a benchmark for multi-step symbolic reasoning, comprising two subsets: MuSR-op (operational reasoning) and MuSR-ta (temporal and abstract reasoning).
% TriviaQA~\cite{joshi2017triviaqa} is a large-scale open-domain QA dataset with web and Wikipedia evidence. 

\subsubsection{Baselines} We compare RAG-FGO with three groups of representative methods. \textbf{Retrieval-based baselines:} 
Direct lets the LLM answer questions without retrieval. 
ReAct~\cite{yao2023react} integrates CoT reasoning with retrieval tool use. 
Self-Act~\cite{press2022measuring} (Self-Ask with Search) decomposes a query into sub-questions and retrieves evidence for each. 
Aflow~\cite{zhang2024aflow} employs an adaptive flow strategy that dynamically decides when to reason or retrieve, making it effective for multi-hop QA. \textbf{Reasoning-based baselines:} 
Zero-shot CoT elicits reasoning by adding a trigger phrase (e.g., ``Let us reason step by step'') to the query without in-context exemplars. Few-shot CoT guides reasoning by providing a small set of annotated exemplars, allowing the model to learn from in-context demonstrations. USC~\cite{chen2023universal} generates multiple CoT reasoning paths and selects the final answer by majority voting. CoT-WP~\cite{wang2024chain} weights reasoning paths by confidence scores from token-level probabilities and aggregates them via weighted voting. LEAP~\cite{zhang2024context} enhances few-shot prompting by inducing errors on exemplars and using self-reflection to extract transferable reasoning principles. ReFeri~\cite{lee2025training} recycles few-shot exemplars to score and verify candidate outputs, selecting responses that are both confident and contextually coherent without additional training. \textbf{Query Rewriting Baselines:} We cover a range of query rewriting baselines, from prompt-driven LLM rewriting to supervised fine-tuning and advanced training paradigms. We include LLM-based methods (LLM Rewrite, HyDE~\cite{gao2023precise}, Query2Doc~\cite{jagerman2023query}, and Crafting~\cite{baek2025crafting}) that generate retrieval-oriented query variants via prompting. We also evaluate SFT, which fine-tunes the rewriter on labeled data. In addition, we consider stronger training-based approaches, including RaFe (DPO)~\cite{rafailov2023direct}, RRR (PPO)~\cite{ma2023query}, and Q-PRM~\cite{ye2025q}. These methods optimize query rewriting through preference learning, reward signals, or process-level supervision. All approaches are evaluated under a unified setting on benchmarks that span multi-hop reasoning, ambiguous queries, and open-domain question answering.

\subsubsection{LLM Generators} 
We evaluate the real-world performance of RAG-FGO using a diverse range of proprietary and open-source LLMs as backbones. These include: GPT-4o-mini and GPT-4.1-nano, representing compact yet powerful proprietary models; Qwen-Long and Qwen-Max, designed for long-context reasoning and high-performance tasks; Qwen2.5-7B-Instruct, a versatile open-source model used to assess instruction-following and reasoning capabilities in practical applications; and Gemma2-27B, a large-scale open-source model optimized for both efficiency and accuracy. We utilize these backbone models to test the end-to-end performance of our optimized query-rewriting agents in realistic retrieval-augmented generation scenarios.
\begin{table*}[t]
\centering
\caption{Comparison with retrieval-based baselines across six QA benchmarks using five LLM generators. Metrics are answer-level F1 scores (\%). Best results are highlighted in bold.}
    \scalebox{1.1}{
        \begin{tabular}{clccccccc}
            \toprule
            \textbf{Model} & \textbf{Method} & \textbf{LongBench} & \textbf{HotpotQA} & \textbf{2Wiki} & \textbf{MuSiQue} & \textbf{NQ} & \textbf{SQuAD} & \textbf{AVG} \\
            \midrule
\multirow{5}{*}{GPT-4o-mini}
 & Direct        & 58.1 & 51.1 & 48.9 & 40.1 & 36.4 & 51.2 & 47.6 \\
 & ReAct         & 62.7 & 65.6 & 52.3 & 53.6 & 48.9 & 55.3 & 56.4 \\
 & Self-Act      & 68.9 & 73.1 & 69.2 & 67.3 & 62.4 & 69.5 & 68.4 \\
 & Aflow         & 61.0 & 73.5 & 65.7 & 71.6 & 67.9 & 73.7 & 68.9 \\
\rowcolor{gray!15}
 & \textbf{RAG-FGO}
 & \textbf{71.5} & \textbf{81.5} & \textbf{77.3}
 & \textbf{79.4} & \textbf{73.8} & \textbf{79.2}
 & \textbf{77.1} \\
\cmidrule(lr){1-9}

\multirow{5}{*}{GPT-4.1-nano}
 & Direct        & 56.3 & 48.5 & 45.3 & 38.2 & 33.7 & 49.3 & 45.2 \\
 & ReAct         & 59.7 & 61.3 & 48.9 & 47.3 & 45.4 & 50.9 & 52.2 \\
 & Self-Act      & 64.5 & 69.7 & 62.2 & 61.1 & 59.8 & 62.7 & 63.3 \\
 & Aflow         & 66.1 & 74.2 & 70.1 & 69.4 & 66.5 & 72.1 & 69.7 \\
\rowcolor{gray!15}
& \textbf{RAG-FGO}
                  & \textbf{67.8} & \textbf{77.8} & \textbf{75.7} & \textbf{73.1} & \textbf{70.3} & \textbf{77.6} & \textbf{73.7} \\
\cmidrule(lr){1-9}

\multirow{5}{*}{Qwen-Long}
 & Direct        & 59.3 & 55.7 & 51.9 & 55.4 & 43.4 & 54.3 & 53.3 \\
 & ReAct         & 64.2 & 69.2 & 57.2 & 58.9 & 56.3 & 58.7 & 60.8 \\
 & Self-Act      & 68.3 & 76.4 & 73.1 & 72.3 & 68.9 & 70.4 & 71.6 \\
 & Aflow         & 70.8 & 80.1 & 76.8 & 78.9 & 74.6 & 77.1 & 76.4 \\
\rowcolor{gray!15}
& \textbf{RAG-FGO}
                  & \textbf{72.3} & \textbf{84.3} & \textbf{80.2} & \textbf{82.6} & \textbf{79.1} & \textbf{80.4} & \textbf{79.8} \\
\cmidrule(lr){1-9}

\multirow{5}{*}{Qwen-Max}
& Direct        & 47.2 & 52.7 & 46.6 & 49.9 & 53.3 & 48.5 & 49.7 \\
& ReAct         & 52.3 & 60.1 & 53.1 & 53.6 & 59.8 & 54.5 & 55.6 \\
& Self-Act      & 58.7 & 65.6 & 64.8 & 61.8 & 67.8 & 68.3 & 64.5 \\
& Aflow         & 61.2 & 71.4 & 68.1 & 65.3 & 73.9 & 74.0 & 69.0 \\
\rowcolor{gray!15}
& \textbf{RAG-FGO}
& \textbf{63.9} & \textbf{76.6} & \textbf{71.1}
& \textbf{67.4} & \textbf{77.3} & \textbf{78.2}
& \textbf{72.4} \\
\cmidrule(lr){1-9}

\multirow{5}{*}{Gemma-2-27B}
& Direct        & 49.3 & 49.7 & 43.4 & 48.4 & 51.3 & 48.5 & 48.4 \\
& ReAct         & 52.5 & 56.3 & 50.8 & 51.7 & 57.8 & 54.5 & 53.9 \\
& Self-Act      & 56.4 & 60.5 & 59.7 & 59.7 & 65.3 & 65.7 & 61.2 \\
& Aflow         & 60.8 & 69.1 & 63.2 & 63.4 & 69.4 & 71.9 & 66.3 \\
\rowcolor{gray!15}
& \textbf{RAG-FGO}
& \textbf{64.9} & \textbf{74.5} & \textbf{66.7}
& \textbf{66.1} & \textbf{73.1} & \textbf{76.6}
& \textbf{70.3} \\
\bottomrule
\end{tabular}
}
\label{tab:model_comparison}
\end{table*}

\begin{table*}[t]
\centering
\caption{Comparison of RAG-FGO with query rewriting baselines on QA benchmarks using Qwen2.5-7B-Instruct as the generator LLM. Best results are highlighted in bold.}
\label{tab:qwen25_7b_results}

\renewcommand{\arraystretch}{1}
\setlength{\tabcolsep}{3pt}

\begin{tabularx}{0.90\textwidth}{l *{12}{>{\centering\arraybackslash}X}}
\toprule
\textbf{Method} 
& \multicolumn{3}{c}{\textbf{HotpotQA}} 
& \multicolumn{3}{c}{\textbf{StrategyQA}} 
& \multicolumn{3}{c}{\textbf{AmbigQA}} 
& \multicolumn{3}{c}{\textbf{PopQA}} \\
\cmidrule(lr){2-4} \cmidrule(lr){5-7} \cmidrule(lr){8-10} \cmidrule(lr){11-13}
& EM & F1 & Acc 
& EM & F1 & Acc 
& EM & F1 & Acc 
& EM & F1 & Acc \\
\midrule

LLM Rewrite & 25.8 & 34.3 & 30.6 & 61.4 & 61.4 & 65.4 & 33.4 & 44.0 & 48.6 & 9.2 & 23.2 & 52.8 \\
HyDE        & 32.8 & 43.8 & 38.6 & 67.2 & 67.2 & 71.0 & 47.2 & 60.4 & 64.6 & 9.0 & 22.7 & 52.8 \\
Query2Doc   & 30.8 & 41.3 & 36.8 & 63.6 & 63.6 & 67.6 & 48.6 & 60.8 & 65.2 & 8.2 & 19.5 & 44.2 \\
Crafting    & 28.8 & 39.3 & 35.4 & 65.2 & 65.2 & 69.0 & 42.8 & 55.1 & 60.8 & 7.6 & 18.4 & 44.8 \\
SFT         & 29.8 & 39.4 & 36.2 & 63.6 & 63.6 & 67.6 & 43.2 & 56.0 & 62.2 & 9.0 & 22.3 & 54.0 \\
RaFe (DPO)  & 32.6 & 43.2 & 38.4 & 68.8 & 68.8 & 71.8 & 48.0 & 60.8 & 64.4 & 11.4 & 26.7 & 58.0 \\
RRR (PPO)   & 33.6 & 44.2 & 39.2 & 68.6 & 68.6 & 71.2 & 47.0 & 59.9 & 64.4 & 12.2 & 28.9 & 62.6 \\
Q-PRM       & 34.2 & 44.5 & 39.4 & 70.2 & 70.2 & 72.6 & 51.0 & 63.9 & 69.2 & 12.2 & 29.8 & 63.6 \\

\midrule
\rowcolor{gray!15}
\textbf{RAG-FGO} 
& \textbf{35.4} & \textbf{45.8} & \textbf{40.1}
& \textbf{72.1} & \textbf{72.1} & \textbf{75.2}
& \textbf{53.2} & \textbf{66.4} & \textbf{71.1}
& \textbf{14.6} & \textbf{31.1} & \textbf{65.2} \\

\bottomrule
\end{tabularx}
\end{table*}
% \begin{table}[t]
% \centering
% \caption{Comparison with classical RAG systems on NQ and TriviaQA (EM scores, \%). RAG-FGO results are mean $\pm$ std over three runs. Best in bold.}
% \scalebox{1.2}{
% \begin{tabular}{lcc}
% \toprule
% \textbf{Method} & \textbf{NQ} & \textbf{TriviaQA} \\
% \midrule
% RAG~\cite{lewis2020retrieval} & 44.5 & 56.8 \\
% FiD-base~\cite{izacard2021leveraging} & 48.2 & 65.0 \\
% FiD-large~\cite{izacard2021leveraging} & 51.4 & 67.6 \\
% RE-RAG-base~\cite{kim2024re} & 49.9 & 68.2 \\
% RE-RAG-large~\cite{kim2024re} & 54.0 & 70.2 \\
% \midrule
% \rowcolor{gray!15} \textbf{RAG-FGO} 
% & \textbf{58.4} & \textbf{73.0} \\
% \bottomrule
% \end{tabular}
% }
% \label{tab:trad_rag_comparison}
% \end{table}
\begin{table*}[t]
\centering
\caption{Comparison of RAG-FGO with baseline methods on seven reasoning benchmarks using GPT-4o-mini as the generator LLM. Best results are highlighted in bold.}
\label{tab:ragfgo_refiri_gpt4omini}
\scalebox{1.15}{
\begin{tabular}{lcccccccc}
\toprule
\textbf{Method} 
 & \textbf{MATH500} & \textbf{MMLU-pro} & \textbf{GPQA} & \textbf{DROP} & \textbf{HotpotQA} & \textbf{MuSR-op} & \textbf{MuSR-ta} & \textbf{Avg.} \\
 & (Acc.) & (Acc.) & (Acc.) & (EM/F1) & (EM) & (Acc.) & (Acc.) &  \\
\midrule
Zero-shot CoT   & 76.4 & 63.0 & 43.0 & 77.6/85.6 & 31.5 & 58.1 & 56.2 & 58.0 \\
Few-shot CoT    & 75.2 & 63.0 & 41.3 & 76.8/83.1 & 34.0 & 59.4 & 77.0 & 61.0 \\
USC             & 78.6 & 62.5 & 42.4 & 78.8/85.8 & 36.6 & 59.8 & 76.4 & 62.2 \\
CoT-WP          & 77.8 & 64.2 & 42.4 & 77.6/83.1 & 34.0 & 57.0 & 79.6 & 61.8 \\
LEAP            & 74.5 & 63.2 & 43.9 & 75.8/83.0 & 34.0 & 59.8 & 74.4 & 60.8 \\
ReFeri  
                & \textbf{78.2} & \textbf{65.0} & 42.4 & \textbf{79.6/85.3} & \textbf{36.2} & 61.3 & 82.8 & 63.6 \\
\rowcolor{gray!15} \textbf{RAG-FGO}  
                & 78.0 & 64.5 & \textbf{44.1} & 79.2/85.0 & 35.9 & \textbf{61.7} & \textbf{83.1} & \textbf{63.7} \\
\bottomrule
\end{tabular}
}
\end{table*}

\begin{table}[t!]
\centering
\caption{Comparison with gradient-free prompt optimization baselines. All results are reported in F1 (\%).}
\label{tab:gradientfree-comparison}
\scalebox{0.9}{
\begin{tabular}{lcccc}
    \toprule
    \textbf{Method} & \textbf{HotpotQA} & \textbf{MuSiQue} & \textbf{NQ} & \textbf{Avg.} \\
    \midrule
    AutoPrompt~\cite{shin2020autoprompt}            & 60.2 & 54.5 & 61.3 &  58.6\\
    Genetic Prompt Search~\cite{xu2022gps}         & 58.7 & 63.9 & 67.9 &  63.5\\
    RLPrompt~\cite{deng2022rlprompt}              & 65.1 & 59.2 & 65.4 &  63.2\\
    \midrule
    \rowcolor{gray!15} \textbf{RAG-FGO}                          
                         & \textbf{74.5} & \textbf{66.1} & \textbf{73.1} & \textbf{71.2} \\
    \bottomrule
\end{tabular}
}
\end{table}

\subsubsection{Implementation Details}
We implement the RAG pipeline with dense retrieval using the E5-base-v2 model~\cite{wang2022text} over the English Wikipedia corpus from the DPR split (December 20, 2018 dump, 21 million 100-word passages)~\cite{karpukhin2020dense}. Pre-processed FAISS indices are sourced from public repositories. Retrieval fetches the top-10 passages ($k=10$) per rewritten query $\hat{q}$ using FAISS~\cite{johnson2019billion} with Inner Product similarity. The query-rewriting agent is optimized using Qwen2.5-7B-Instruct as the rewriter LLM, while final answer generation is performed by the backbone LLMs.

To maintain consistent performance across varying levels of task complexity, we employ hyperparameter configurations that scale with the retrieval difficulty of each benchmark (e.g., $mc=9$ for LongBench and $mc=12$ for HotpotQA). We further set a constant elite-preserving memory pool capacity of $M=5$ for all experiments. These settings represent balanced search budgets that are further justified by the sensitivity analysis in our ablation study. Unless otherwise specified, all reported metrics are based on the final optimized agents obtained at the conclusion of the RAG-FGO training process.

\subsection{Performance Evaluation}
\subsubsection{Retrieval-based baselines}
We compare RAG-FGO with four representative retrieval-based baselines across six QA benchmarks and five LLM backbones, as summarized in Table~\ref{tab:model_comparison}. RAG-FGO consistently achieves the strongest performance across all backbones, indicating that its advantages are not tied to a specific model family.

On the GPT-4o-mini backbone, RAG-FGO substantially outperforms competitive baselines such as Aflow and Self-Act on retrieval-intensive benchmarks, including HotpotQA, 2Wiki, and MuSiQue. These gains suggest that the proposed search-based optimization effectively aligns rewritten queries with multi-hop and compositional evidence requirements. Notably, RAG-FGO also yields consistent improvements on knowledge-heavy datasets such as NQ and SQuAD, demonstrating that the optimized rewriting strategy generalizes beyond explicit multi-hop settings. Similar patterns are observed across other backbones, including Qwen-Long, Qwen-Max, and Gemma-2-27B, where RAG-FGO maintains clear margins over fixed-prompt agentic workflows. Overall, these results highlight that strategy-level refinement of query rewriting provides a more robust foundation for retrieval-augmented generation than static or heuristic-based baselines.

\subsubsection{Query-Rewriting Baselines}
Table ~\ref{tab:qwen25_7b_results} compares RAG-FGO with representative query rewriting baselines under a unified Qwen2.5-7B-Instruct backbone.
Across all four benchmarks, RAG-FGO consistently achieves the best performance on EM, F1, and accuracy, indicating a robust improvement rather than isolated gains on specific datasets. Notably, the improvements are most pronounced on AmbigQA and PopQA, where semantic ambiguity and long-tail knowledge pose significant challenges to retrieval. This suggests that the iterative, feedback-driven optimization in RAG-FGO effectively reduces query–document mismatch in scenarios where naive rewriting strategies are insufficient.
In contrast, gains on HotpotQA and StrategyQA, while more moderate, remain consistent across all metrics, demonstrating that RAG-FGO enhances evidence retrieval without disrupting multi-hop or implicit reasoning.

Overall, these results highlight that RAG-FGO provides a generalizable optimization of query rewriting strategies, yielding stable improvements across diverse QA settings with different reasoning and retrieval characteristics.

\subsubsection{Reasoning-Centric Baselines}
As shown in Table~\ref{tab:ragfgo_refiri_gpt4omini}, RAG-FGO achieves the best overall average score (63.7) on the seven reasoning benchmarks, narrowly surpassing ReFeri (63.6). Beyond the average, RAG-FGO shows its clearest benefits on knowledge-intensive and multi-hop benchmarks such as GPQA and MuSR, where optimized rewriting better aligns retrieval with the underlying evidence needs, yielding consistent gains over strong prompt-based baselines.
\begin{table}[t!]
\centering
\caption{Performance comparison of different heuristic optimizers under identical RAG configurations, using GPT-4o-mini as the generator LLM. Results are reported in F1 (\%).}
\label{tab:heuristic_comparison}
\begin{tabular}{lccc}
\toprule
\textbf{Method} & \textbf{LongBench} & \textbf{HotpotQA} & \textbf{2WikiMultihopQA} \\
\midrule
PSO~\cite{kennedy1995particle}      & 66.2 & 72.9 & 69.1 \\
GA~\cite{lambora2019genetic}                & 65.5 & 71.3 & 67.8 \\
SA ~\cite{kirkpatrick1983optimization}               & 61.0 & 67.2 & 63.5 \\
FOA ~\cite{ghaemi2014forest}    & 68.8 & 76.4 & 72.5 \\
TLBO~\cite{rao2011teaching} & 63.3 & 69.0 & 65.4 \\
HS~\cite{geem2001new}                     & 64.1 & 70.2 & 66.7 \\
\midrule
\rowcolor{gray!15} \textbf{RAG-FGO}                & \textbf{71.5} & \textbf{81.5} & \textbf{77.3} \\
\bottomrule
\end{tabular}
\end{table}
On a few benchmarks that are comparatively less constrained by external evidence (e.g., MATH500 and MMLU-pro), the improvements are smaller and RAG-FGO can be slightly behind the strongest prompt-based method. Overall, these results indicate that RAG-FGO is most effective when solving requires grounding in retrieved knowledge, while remaining competitive on reasoning-centric tasks with weaker reliance on retrieval.

\subsubsection{Experimental Comparison with Prompt Optimization Baselines}
To situate RAG-FGO within the broader space of gradient-free and evolutionary 
prompt optimization techniques, we compare it against AutoPrompt, Genetic Prompt 
Search, and RLPrompt under identical RAG configurations. All methods use the same 
retriever and the Gemma-2-27B generator to ensure a controlled and comparable 
evaluation setting. Because these existing approaches were originally developed 
for classification or masked-LM scenarios rather than retrieval-augmented generation 
with decoder-only models, we implement adapted, gradient-free variants suitable 
for query-rewriting prompts. As shown in Table~\ref{tab:gradientfree-comparison}, 
RAG-FGO achieves higher scores across HotpotQA, MuSiQue, and NQ. These results 
suggest that the performance differences arise from the structured global--local 
search procedure introduced in RAG-FGO rather than from generic prompt perturbation 
or search diversity alone.

\subsubsection{Evaluation Against Heuristic Optimization Paradigms}
We further examine RAG-FGO in comparison with several widely used heuristic optimization paradigms, including PSO, GA, SA, FOA, TLBO, and HS, under the same RAG configuration with GPT-4o-mini. 
Rather than treating these methods as specialized optimizers for prompt agents, we apply them in a direct manner to represent different heuristic search philosophies commonly used in black-box optimization.

As shown in Table~\ref{tab:heuristic_comparison}, FOA yields the strongest performance among the heuristic paradigms, while PSO and GA achieve moderate results, and SA, TLBO, and HS perform less competitively across datasets. 
RAG-FGO consistently outperforms all heuristic baselines on the three benchmarks, suggesting that a structured combination of global search and local search, tailored to the RAG agent setting, is more effective than directly transferring generic heuristic search strategies.

\subsubsection{Cross-Task Generalization Experiments}
Transfer results across QA benchmarks suggest that the query-rewriting agent optimized on HotpotQA can generalize to other tasks without further tuning while maintaining relatively strong performance (Table~\ref{tab:hotpot_to_singlehop}). Although performance drops are observed across all targets, the degradation is more pronounced on AmbigQA and StrategyQA, where the question formulations and supervision signals differ more substantially from those of HotpotQA.

\begin{table}[t]
\centering
\caption{Cross-task generalization from HotpotQA using GPT-4o-mini as the generator LLM. Target-tuned: performance with target-specific optimization; HotpotQA$\rightarrow$Target: performance with the agent transferred from HotpotQA; $\Delta$: drop in F1 score. Results are reported in F1 (\%)}
\label{tab:hotpot_to_singlehop}
\begin{tabular}{lccc}
\toprule
\textbf{Target Dataset} & \textbf{Target-tuned} & \textbf{HotpotQA$\rightarrow$Target} & \textbf{$\Delta$} \\
\midrule
NQ                      & 73.8                  & 71.1                                 & -2.7 \\
SQuAD                   & 79.2                  & 77.3                                 & -1.9 \\
AmbigQA                 & 63.9                  & 59.8                                 & -4.1 \\
StrategyQA              & 70.2                  & 64.5                                 & -5.7 \\
\bottomrule
\end{tabular}
\end{table}
In particular, AmbigQA emphasizes ambiguity resolution across multiple plausible answers, and StrategyQA focuses on short, implicit reasoning with limited reliance on explicit multi-hop evidence. As a result, a rewriting policy optimized for multi-hop evidence aggregation may not be fully aligned with the retrieval patterns favored by these benchmarks. Nevertheless, the transferred agent remains competitive across tasks, suggesting that RAG-FGO captures a rewriting behavior that is partially robust to task shifts. At the same time, the observed gaps also confirm that task-specific optimization continues to provide additional benefits when the target task departs from the characteristics of the source dataset.

\subsection{Ablation Study}
\label{sec:Ablation}
\begin{figure*}[t]
    \centering
    \includegraphics[width=1\textwidth]{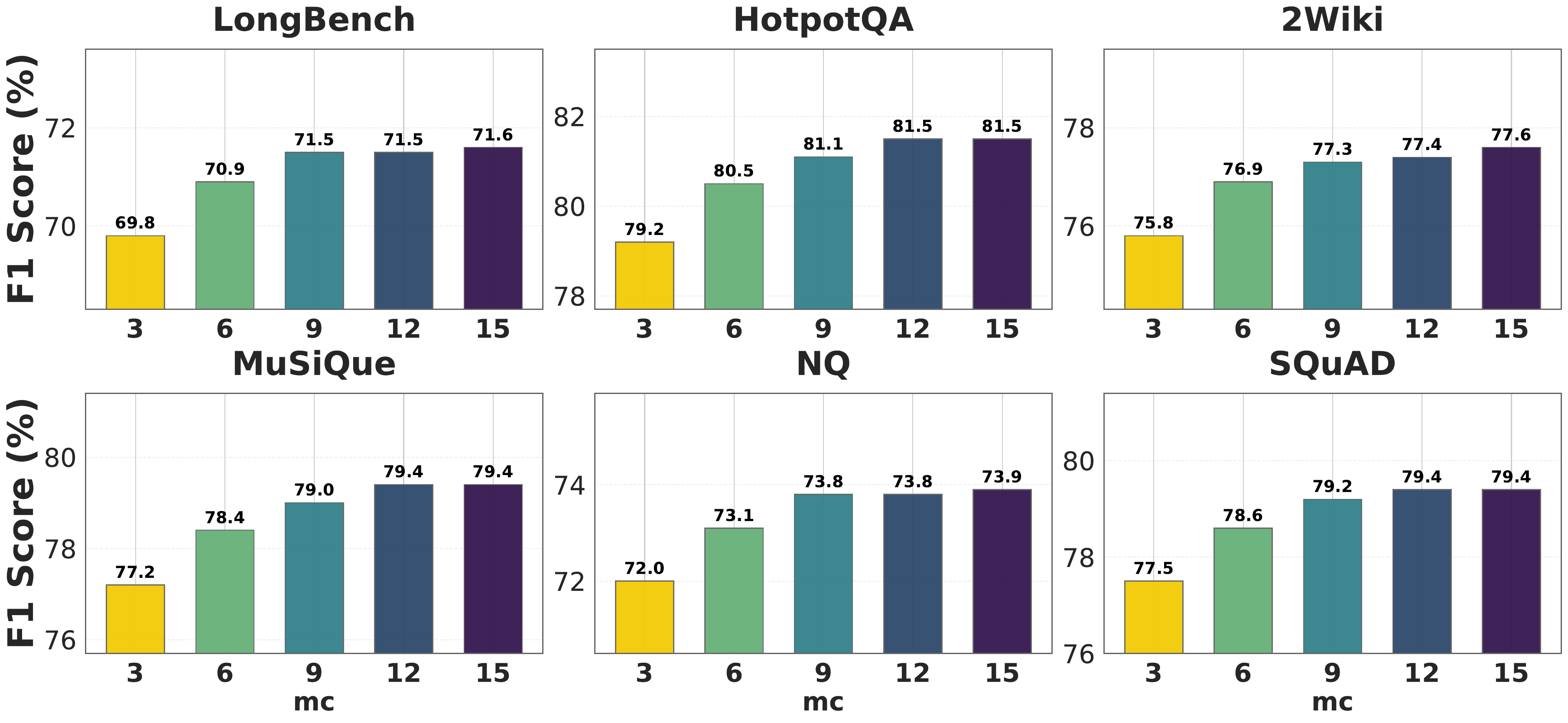}
    \caption{Performance (F1, \%) as a function of mutation count ($  mc$) on six QA benchmarks using GPT-4o-mini.}
    \label{fig:experiment01}
\end{figure*}

\subsubsection{Impact of mc on Performance Saturation and Token Consumption} 
\begin{figure}[t]
    \centering
    \includegraphics[width=1\linewidth]{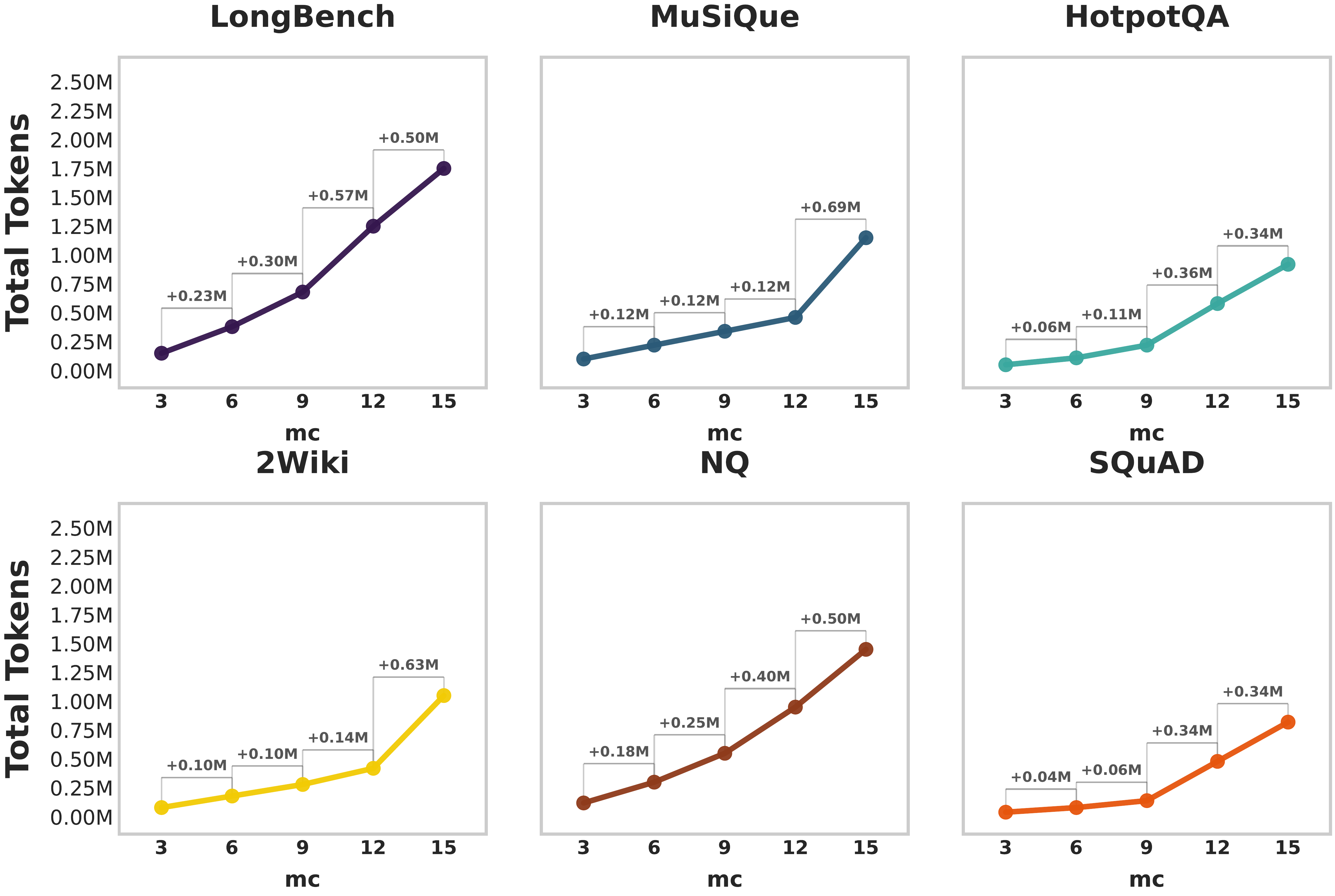}
    \caption{Total token consumption under different $mc$ settings across six QA datasets. Each subplot corresponds to a dataset. The x-axis denotes the mutation count $mc$, and the y-axis reports the total tokens (in millions, M).}
    \label{fig:experiment02}
\end{figure}

We further investigate the impact of the mutation count ($mc$) on both performance and computational cost across the six QA benchmarks using GPT-4o-mini as the generator LLM (Figures~\ref{fig:experiment01} and~\ref{fig:experiment02}). As shown in Figure ~\ref{fig:experiment01}, F1 scores generally increase as $mc$ grows from 3 to 15 across all datasets. The gains are most pronounced when moving from $mc=3$ to $mc=6$ or $mc=9$, after which performance improvements become marginal or saturate (e.g., LongBench plateaus around 71.5–71.6\%, HotpotQA around 81.5\%, and SQuAD at 79.4\% beyond $mc=9$). This saturation pattern suggests that increasing the number of candidate mutations beyond a moderate range yields diminishing returns in terms of final answer quality.
Figure~\ref{fig:experiment02} complements this observation by illustrating the corresponding growth in total token consumption during the optimization process. Token usage increases rapidly with $mc$, with particularly sharp rises at higher values (e.g., from $mc=9$ to $mc=15$, the additional cost ranges from +0.34M tokens on SQuAD/HotpotQA to +0.50M–0.69M on LongBench, MuSiQue, and NQ). Taken together, the two figures reveal a clear accuracy–cost trade-off: configurations around $mc=9$ to $mc=12$ (as adopted in our main experiments) achieve near-saturation performance while keeping token expenditure within a reasonable range (typically 0.5–1.25M tokens across datasets), whereas pushing $mc$ to 15 provides only marginal F1 gains (0–0.6\%) at disproportionately higher cost. These results confirm that moderate $mc$ values (around 9–12, depending on the dataset) represent an effective sweet spot for RAG-FGO, balancing retrieval-augmented performance and optimization efficiency.

\subsubsection{Component Ablation}
Table~\ref{tab:component_ablation_multi_dataset} indicates that each module of RAG-FGO contributes meaningfully, but their importance differs. Removing the elite-preserving memory pool leads to the largest degradation, with roughly 6--7 F1 drop across all datasets. This consistent gap suggests that retaining high-performing agents throughout the search process is crucial: without the elite-preserving memory pool, the optimization trajectory becomes more volatile, and strong candidates are more likely to be overwritten by short-term fluctuations, resulting in a noticeably weaker final agent.
Removing local search causes a moderate but stable decrease (around 6--7\% relative drop on average), highlighting the value of fine-grained refinements around strong candidates. Disabling global search yields a smaller drop (around 4--5\% relative drop), suggesting that broad exploration mainly improves early-stage coverage, while later-stage gains depend more on refinement and preservation of effective strategies. Overall, the ablation results support a complementary design: global search provides diversity, local search improves precision, and the elite-preserving memory pool stabilizes progress by retaining strong agents across iterations. We next provide a process-level analysis to explain why the elite-preserving memory pool yields the largest gains.
\begin{figure}[t]
    \centering
    \includegraphics[width=0.9\linewidth]{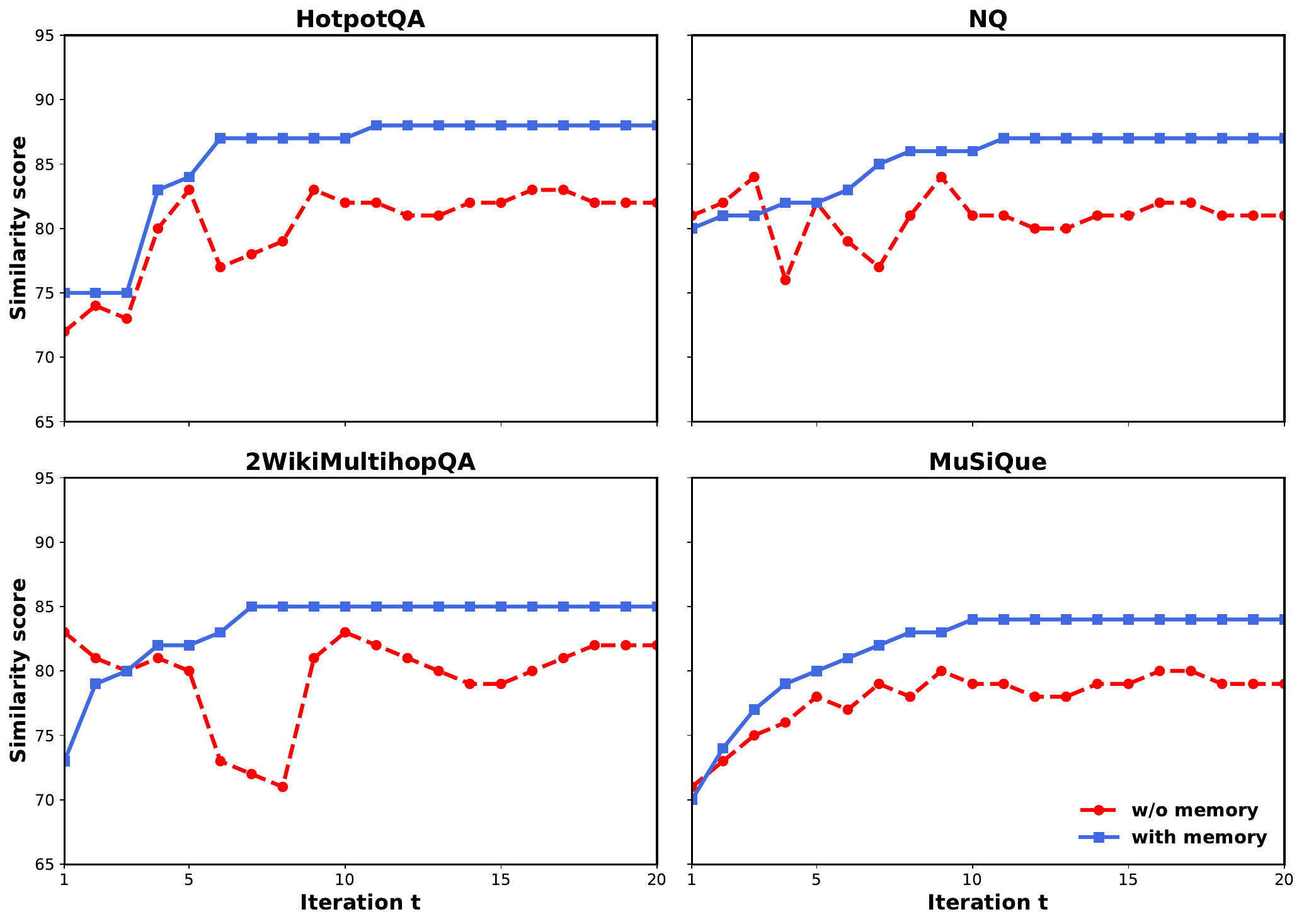}
    \caption{Effect of elite-preserving memory pool on saturation during RAG-FGO optimization process.}
    \label{fig:placeholder}
\end{figure}

\begin{table}[t] 
\centering
\caption{Component ablation of RAG-FGO across datasets (GPT-4o-mini). We report answer-level F1 (\%), and the relative change w.r.t. the full model in parentheses.}
\label{tab:component_ablation_multi_dataset}
\resizebox{\columnwidth}{!}{
\begin{tabular}{lccccc}
\toprule
\textbf{Configuration} & \textbf{Hotpot} & \textbf{MuSiQue} & \textbf{2Wiki} & \textbf{NQ} & \textbf{SQuAD} \\
\midrule
\rowcolor{gray!15} \textbf{RAG-FGO (full)} & \textbf{81.5} & \textbf{79.4} & \textbf{77.3} & \textbf{73.8} & \textbf{79.2} \\
w/o Memory Pool & 75.2 (-6.3) & 72.8 (-6.6) & 70.9 (-6.4) & 66.9 (-6.9) & 72.1 (-7.1) \\
w/o Local Search & 76.4 (-5.1) & 74.2 (-5.2) & 72.1 (-5.2) & 69.0 (-4.8) & 74.3 (-4.9) \\
w/o Global Search & 78.3 (-3.2) & 75.4 (-4.0) & 74.0 (-3.3) & 71.2 (-2.6) & 76.1 (-3.1) \\
\bottomrule
\end{tabular}
}
\end{table}

\subsubsection{Effect of elite-preserving Memory Pool on Optimization Dynamics}

Figure~\ref{fig:placeholder} compares the semantic-similarity trajectories during training with and without the elite-preserving memory pool. Across all datasets, the variant with memory exhibits a faster rise in similarity and reaches an earlier plateau, whereas the variant without memory shows larger oscillations and a lower stable level. This pattern is especially clear on HotpotQA and 2WikiMultihopQA, where the no-memory setting undergoes pronounced fluctuations before settling, while the memory-based variant quickly stabilizes at a higher similarity score. These trajectories offer a direct explanation for the large performance drop observed in Table~\ref{tab:component_ablation_multi_dataset} when the elite-preserving memory pool is removed. By retaining the best-performing agents across iterations, the elite-preserving memory pool prevents effective strategies from being discarded due to transient noise in the evaluation signal, leading to more stable optimization behavior and stronger final agents.

\subsubsection{Hyperparameter Sensitivity Analysis}
To evaluate the robustness of RAG-FGO with respect to its search-related hyperparameters, we conduct a one-factor-at-a-time sensitivity analysis on HotpotQA. We examine five parameters that control the behavior of agent evolution: the minimum-change threshold $\delta$ used in global perturbations, the local and global perturbation scales $(\sigma_l, \sigma_g)$, and the corresponding range control coefficients $(k_l, k_g)$. Each parameter is varied independently within a practical neighborhood around the default configuration, while all other settings are kept fixed.

For each setting, we report the relative change in answer-level F1 on HotpotQA and the total token cost incurred during optimization, normalized with respect to the default configuration. The reported ranges reflect variability across multiple independent runs (Table~\ref{tab:param_sensitivity}). Overall, the results indicate that RAG-FGO is relatively insensitive to moderate variations of these hyperparameters, with performance changes typically within $\pm1\%$ in a broad neighborhood of the default setting. Notably, more aggressive exploration settings (e.g., larger $\sigma_g$ or $k_g$) can yield marginal F1 improvements, but at a substantially higher token cost. Conversely, overly conservative configurations reduce both cost and performance. These trends suggest that the default hyperparameter values are chosen as a balanced trade-off between retrieval effectiveness and optimization efficiency, rather than as per-dataset or per-task optimal settings.

\begin{table}[t!]
\centering
\caption{Sensitivity analysis of search-related hyperparameters $\delta,\sigma_l,\sigma_g,k_l,k_g$ on HotpotQA. 
We report normalized changes in answer-level F1 and optimization cost (token usage) relative to the default configuration 
($\delta = 1.00,\ \sigma_l = 1.00,\ \sigma_g = 0.50,\ k_l = 0.50,\ k_g = 0.10$).
}
\label{tab:param_sensitivity}
\scriptsize
\setlength{\tabcolsep}{8pt}
\renewcommand{\arraystretch}{0.9}
\begin{tabular}{clcc}
\toprule
\textbf{Param} & \textbf{Value} & \textbf{Avg F1} & \textbf{Cost} \\
\midrule
\multirow{5}{*}{$\delta$}
 & 0.50 & ↓\,0.8--1.5\% & ↓\,2--4\% \\
 & 0.75 & ↓\,0.1--0.5\% & ↓\,1--2\% \\
 & 1.00 & baseline      & baseline \\
 & 1.25 & ↑\,0.3--0.9\% & ↑\,1--3\% \\
 & 1.50 & ↓\,0.5--1.2\% & ↑\,1--3\% \\
\midrule
\multirow{5}{*}{$\sigma_l$}
 & 0.50 & ↓\,0.6--1.2\% & ↓\,2--4\% \\
 & 0.75 & ↓\,0.1--0.6\% & ↓\,1--2\% \\
 & 1.00 & baseline      & baseline \\
 & 1.25 & ↑\,0.3--0.9\% & ↑\,1--3\% \\
 & 1.50 & ↓\,0.5--1.3\% & ↑\,1--3\% \\
\midrule
\multirow{5}{*}{$\sigma_g$}
 & 0.00 & ↓\,1.0--1.5\% & ↓\,4--7\% \\
 & 0.25 & ↓\,0.5--1.0\% & ↓\,2--4\% \\
 & 0.50 & baseline      & baseline \\
 & 0.75 & ↑\,0.2--0.8\% & ↑\,1--3\% \\
 & 1.00 & ↑\,0.8--1.5\% & ↑\,3--5\% \\
\midrule
\multirow{5}{*}{$k_l$}
 & 0.00 & ↓\,1.0--1.5\% & ↓\,3--6\% \\
 & 0.25 & ↓\,0.4--0.9\% & ↓\,2--4\% \\
 & 0.50 & baseline      & baseline \\
 & 0.75 & ↑\,0.3--0.9\% & ↑\,1--3\% \\
 & 1.00 & ↓\,0.5--1.3\% & ↑\,1--3\% \\
\midrule
\multirow{4}{*}{$k_g$}
 & 0.05 & ↓\,0.1--0.6\% & ↓\,1--3\% \\
 & 0.10 & baseline      & baseline \\
 & 0.20 & ↑\,0.3--1.0\% & ↑\,2--5\% \\  
 & 0.30 & ↓\,0.5--1.5\% & ↑\,4--8\% \\
\bottomrule
\end{tabular}
\end{table}

\section{Conclusion}
In this work, we present RAG-FGO, a dynamic optimization framework that reframes generative query rewriting in retrieval-augmented generation as a joint optimization problem over discrete prompt templates and continuous decoding parameters. By integrating fungal growth-inspired exploration–exploitation with an elite-preserving memory pool, RAG-FGO effectively navigates noisy, non-stationary, mixed discrete–continuous strategy spaces in real-world RAG systems. This study highlights two core lessons: treating query reformulation as an iteratively learnable agent rather than a fixed preprocessing step unlocks substantial gains in retrieval quality under complex semantics; and balancing global semantic jumps with local refinements, stabilized by historical elite retention, is essential to overcome stochastic feedback and escape suboptimal local optima. Looking forward, promising directions include lightweight online adaptation for continuous agent evolution during deployment, extension to multi-modal knowledge bases and continual learning under distribution shifts, and more efficient surrogate objectives or meta-optimization to reduce computational overhead while retaining robustness.

\section{Limitation}
Despite its effectiveness, the proposed RAG-FGO framework has several limitations. The optimization procedure is conducted in an offline stage that requires an additional training budget prior to deployment, which may reduce flexibility in scenarios involving rapid iteration or frequent updates. Although the optimized query-rewriting agent operates efficiently at inference time with a single forward pass, the current framework does not support online or continual adaptation to evolving queries or dynamically changing knowledge bases. 
Future work will explore lightweight online adaptation strategies and more efficient optimization schemes, as well as further evaluation in domain-specific knowledge systems under noisy black-box feedback.

\section*{Acknowledgments}
This work was supported by the National Natural Science Foundation of China (NSFC) under the General Program (Grant No. 62572104).

This article used large language models (such as ChatGPT) as an auxiliary tool in the language polishing process, but did not use them in research conception and academic content generation.

\bibliographystyle{IEEEtran}
\bibliography{reference}

@article{bai2023longbench,
  title={Longbench: A bilingual, multitask benchmark for long context understanding},
  author={Bai, Yushi and Lv, Xin and Zhang, Jiajie and Lyu, Hongchang and Tang, Jiankai and Huang, Zhidian and Du, Zhengxiao and Liu, Xiao and Zeng, Aohan and Hou, Lei and others},
  journal={arXiv preprint arXiv:2308.14508},
  year={2023}
}

@article{rajpurkar2016squad,
  title={Squad: 100,000+ questions for machine comprehension of text},
  author={Rajpurkar, Pranav and Zhang, Jian and Lopyrev, Konstantin and Liang, Percy},
  journal={arXiv preprint arXiv:1606.05250},
  year={2016}
}

@article{yang2018hotpotqa,
  title={HotpotQA: A dataset for diverse, explainable multi-hop question answering},
  author={Yang, Zhilin and Qi, Peng and Zhang, Saizheng and Bengio, Yoshua and Cohen, William W and Salakhutdinov, Ruslan and Manning, Christopher D},
  journal={arXiv preprint arXiv:1809.09600},
  year={2018}
}

@article{kwiatkowski2019natural,
  title={Natural questions: a benchmark for question answering research},
  author={Kwiatkowski, Tom and Palomaki, Jennimaria and Redfield, Olivia and Collins, Michael and Parikh, Ankur and Alberti, Chris and Epstein, Danielle and Polosukhin, Illia and Devlin, Jacob and Lee, Kenton and others},
  journal={Transactions of the Association for Computational Linguistics},
  volume={7},
  pages={453--466},
  year={2019},
  publisher={MIT Press One Rogers Street, Cambridge, MA 02142-1209, USA journals-info~…}
}

@article{trivedi2022musique,
  title={MuSiQue: Multihop Questions via Single-hop Question Composition},
  author={Trivedi, Harsh and Balasubramanian, Niranjan and Khot, Tushar and Sabharwal, Ashish},
  journal={Transactions of the Association for Computational Linguistics},
  volume={10},
  pages={539--554},
  year={2022},
  publisher={MIT Press One Broadway, 12th Floor, Cambridge, Massachusetts 02142, USA~…}
}

@article{ho2020constructing,
  title={Constructing a multi-hop qa dataset for comprehensive evaluation of reasoning steps},
  author={Ho, Xanh and Nguyen, Anh-Khoa Duong and Sugawara, Saku and Aizawa, Akiko},
  journal={arXiv preprint arXiv:2011.01060},
  year={2020}
}

@article{abdel2025fungal,
  title={Fungal growth optimizer: A novel nature-inspired metaheuristic algorithm for stochastic optimization},
  author={Abdel-Basset, Mohamed and Mohamed, Reda and Abouhawwash, Mohamed},
  journal={Computer Methods in Applied Mechanics and Engineering},
  volume={437},
  pages={117825},
  year={2025},
  publisher={Elsevier}
}

@article{press2022measuring,
  title={Measuring and narrowing the compositionality gap in language models},
  author={Press, Ofir and Zhang, Muru and Min, Sewon and Schmidt, Ludwig and Smith, Noah A and Lewis, Mike},
  journal={arXiv preprint arXiv:2210.03350},
  year={2022}
}

@inproceedings{yao2023react,
  title={React: Synergizing reasoning and acting in language models},
  author={Yao, Shunyu and Zhao, Jeffrey and Yu, Dian and Du, Nan and Shafran, Izhak and Narasimhan, Karthik and Cao, Yuan},
  booktitle={International Conference on Learning Representations (ICLR)},
  year={2023}
}

@inproceedings{gao2023precise,
  title={Precise zero-shot dense retrieval without relevance labels},
  author={Gao, Luyu and Ma, Xueguang and Lin, Jimmy and Callan, Jamie},
  booktitle={Proceedings of the 61st Annual Meeting of the Association for Computational Linguistics (Volume 1: Long Papers)},
  pages={1762--1777},
  year={2023}
}

@inproceedings{qolomany2017parameters,
  title={Parameters optimization of deep learning models using particle swarm optimization},
  author={Qolomany, Basheer and Maabreh, Majdi and Al-Fuqaha, Ala and Gupta, Ajay and Benhaddou, Driss},
  booktitle={2017 13th International Wireless Communications and Mobile Computing Conference (IWCMC)},
  pages={1285--1290},
  year={2017},
  organization={IEEE}
}

@INPROCEEDINGS{10611913,
  author={Liu, Shengcai and Chen, Caishun and Qu, Xinghua and Tang, Ke and Ong, Yew-Soon},
  booktitle={2024 IEEE Congress on Evolutionary Computation (CEC)},
  title={Large Language Models as Evolutionary Optimizers},
  year={2024},
  volume={},
  number={},
  pages={1-8},
  doi={10.1109/CEC60901.2024.10611913}}

@article{kirkpatrick1983optimization,
  title={Optimization by simulated annealing},
  author={Kirkpatrick, Scott and Gelatt Jr, C Daniel and Vecchi, Mario P},
  journal={science},
  volume={220},
  number={4598},
  pages={671--680},
  year={1983},
  publisher={American association for the advancement of science}
}

@book{talbi2009metaheuristics,
  title={Metaheuristics: from design to implementation},
  author={Talbi, El-Ghazali},
  year={2009},
  publisher={John Wiley \& Sons}
}

@article{blum2003metaheuristics,
  title={Metaheuristics in combinatorial optimization: Overview and conceptual comparison},
  author={Blum, Christian and Roli, Andrea},
  journal={ACM computing surveys (CSUR)},
  volume={35},
  number={3},
  pages={268--308},
  year={2003},
  publisher={Acm New York, NY, USA}
}

@book{yang2020nature,
  title={Nature-inspired optimization algorithms},
  author={Yang, Xin-She},
  year={2020},
  publisher={Academic Press}
}

@article{gao2023retrieval,
  title={Retrieval-augmented generation for large language models: A survey},
  author={Gao, Yunfan and Xiong, Yun and Gao, Xinyu and Jia, Kangxiang and Pan, Jinliu and Bi, Yuxi and Dai, Yi and Sun, Jiawei and Wang, Haofen and Wang, Haofen},
  journal={arXiv preprint arXiv:2312.10997},
  volume={2},
  year={2023}
}

@article{shi2023replug,
  title={Replug: Retrieval-augmented black-box language models},
  author={Shi, Weijia and Min, Sewon and Yasunaga, Michihiro and Seo, Minjoon and James, Rich and Lewis, Mike and Zettlemoyer, Luke and Yih, Wen-tau},
  journal={arXiv preprint arXiv:2301.12652},
  year={2023}
}

@article{mao2021rider,
  title={RIDER: Reader-Guided Passage Reranking for Open-Domain Question Answering},
  author={Mao, Yuning and He, Pengcheng and Liu, Xiaodong and Shen, Yelong and Gao, Jianfeng and Han, Jiawei and Chen, Weizhu},
  journal={arXiv preprint arXiv:2101.00294},
  year={2021}
}

@article{mialon2023augmented,
  title={Augmented language models: a survey},
  author={Mialon, Gr{\'e}goire and Dess{\`\i}, Roberto and Lomeli, Maria and Nalmpantis, Christoforos and Pasunuru, Ram and Raileanu, Roberta and Rozi{\`e}re, Baptiste and Schick, Timo and Dwivedi-Yu, Jane and Celikyilmaz, Asli and others},
  journal={arXiv preprint arXiv:2302.07842},
  year={2023}
}

@inproceedings{khattab2020colbert,
  title={Colbert: Efficient and effective passage search via contextualized late interaction over bert},
  author={Khattab, Omar and Zaharia, Matei},
  booktitle={Proceedings of the 43rd International ACM SIGIR conference on research and development in Information Retrieval},
  pages={39--48},
  year={2020}
}

@inproceedings{karpukhin2020dense,
  title={Dense Passage Retrieval for Open-Domain Question Answering.},
  author={Karpukhin, Vladimir and Oguz, Barlas and Min, Sewon and Lewis, Patrick SH and Wu, Ledell and Edunov, Sergey and Chen, Danqi and Yih, Wen-tau},
  booktitle={EMNLP (1)},
  pages={6769--6781},
  year={2020}
}

@article{izacard2020leveraging,
  title={Leveraging passage retrieval with generative models for open domain question answering},
  author={Izacard, Gautier and Grave, Edouard},
  journal={arXiv preprint arXiv:2007.01282},
  year={2020}
}

@article{johnson2019billion,
  title={Billion-scale similarity search with GPUs},
  author={Johnson, Jeff and Douze, Matthijs and J{\'e}gou, Herv{\'e}},
  journal={IEEE Transactions on Big Data},
  volume={7},
  number={3},
  pages={535--547},
  year={2019},
  publisher={IEEE}
}

@inproceedings{jiang2023active,
  title={Active retrieval augmented generation},
  author={Jiang, Zhengbao and Xu, Frank F and Gao, Luyu and Sun, Zhiqing and Liu, Qian and Dwivedi-Yu, Jane and Yang, Yiming and Callan, Jamie and Neubig, Graham},
  booktitle={Proceedings of the 2023 Conference on Empirical Methods in Natural Language Processing},
  pages={7969--7992},
  year={2023}
}

@article{zhang2024aflow,
  title={Aflow: Automating agentic workflow generation},
  author={Zhang, Jiayi and Xiang, Jinyu and Yu, Zhaoyang and Teng, Fengwei and Chen, Xionghui and Chen, Jiaqi and Zhuge, Mingchen and Cheng, Xin and Hong, Sirui and Wang, Jinlin and others},
  journal={arXiv preprint arXiv:2410.10762},
  year={2024}
}

@article{lewis2020retrieval,
  title={Retrieval-Augmented Generation for Knowledge-Intensive NLP Tasks},
  author={Lewis, Patrick and Perez, Ethan and Piktus, Aleksandra and Petroni, Fabio and Karpukhin, Vladimir and Goyal, Naman and Küttler, Heinrich and Lewis, Mike and Yih, Wen-tau and Rocktäschel, Tim and Riedel, Sebastian},
  journal={arXiv preprint arXiv:2005.11401},
  year={2020},
  url={https://arxiv.org/abs/2005.11401}
}

@article{hendrycks2021measuring,
  title={Measuring mathematical problem solving with the math dataset},
  author={Hendrycks, Dan and Burns, Collin and Kadavath, Saurav and Arora, Akul and Basart, Steven and Tang, Eric and Song, Dawn and Steinhardt, Jacob},
  journal={arXiv preprint arXiv:2103.03874},
  year={2021}
}

@article{wang2024mmlu,
  title={Mmlu-pro: A more robust and challenging multi-task language understanding benchmark},
  author={Wang, Yubo and Ma, Xueguang and Zhang, Ge and Ni, Yuansheng and Chandra, Abhranil and Guo, Shiguang and Ren, Weiming and Arulraj, Aaran and He, Xuan and Jiang, Ziyan and others},
  journal={Advances in Neural Information Processing Systems},
  volume={37},
  pages={95266--95290},
  year={2024}
}

@inproceedings{rein2024gpqa,
  title={Gpqa: A graduate-level google-proof q\&a benchmark},
  author={Rein, David and Hou, Betty Li and Stickland, Asa Cooper and Petty, Jackson and Pang, Richard Yuanzhe and Dirani, Julien and Michael, Julian and Bowman, Samuel R},
  booktitle={First Conference on Language Modeling},
  year={2024}
}

@article{dua2019drop,
  title={DROP: A reading comprehension benchmark requiring discrete reasoning over paragraphs},
  author={Dua, Dheeru and Wang, Yizhong and Dasigi, Pradeep and Stanovsky, Gabriel and Singh, Sameer and Gardner, Matt},
  journal={arXiv preprint arXiv:1903.00161},
  year={2019}
}

@article{sprague2023musr,
  title={Musr: Testing the limits of chain-of-thought with multistep soft reasoning},
  author={Sprague, Zayne and Ye, Xi and Bostrom, Kaj and Chaudhuri, Swarat and Durrett, Greg},
  journal={arXiv preprint arXiv:2310.16049},
  year={2023}
}

@article{chen2023universal,
  title={Universal self-consistency for large language model generation},
  author={Chen, Xinyun and Aksitov, Renat and Alon, Uri and Ren, Jie and Xiao, Kefan and Yin, Pengcheng and Prakash, Sushant and Sutton, Charles and Wang, Xuezhi and Zhou, Denny},
  journal={arXiv preprint arXiv:2311.17311},
  year={2023}
}

@article{wang2024chain,
  title={Chain-of-thought reasoning without prompting},
  author={Wang, Xuezhi and Zhou, Denny},
  journal={Advances in Neural Information Processing Systems},
  volume={37},
  pages={66383--66409},
  year={2024}
}

@article{zhang2024context,
  title={In-context principle learning from mistakes},
  author={Zhang, Tianjun and Madaan, Aman and Gao, Luyu and Zheng, Steven and Mishra, Swaroop and Yang, Yiming and Tandon, Niket and Alon, Uri},
  journal={arXiv preprint arXiv:2402.05403},
  year={2024}
}

@article{lee2025training,
  title={Training-free LLM Verification via Recycling Few-shot Examples},
  author={Lee, Dongseok and Hong, Jimyung and Kim, Dongyoung and Kim, Jaehyung},
  journal={arXiv preprint arXiv:2506.17251},
  year={2025}
}

@article{fernando2023promptbreeder,
  title={Promptbreeder: Self-referential self-improvement via prompt evolution},
  author={Fernando, Chrisantha and Banarse, Dylan and Michalewski, Henryk and Osindero, Simon and Rockt{\"a}schel, Tim},
  journal={arXiv preprint arXiv:2309.16797},
  year={2023}
}

@inproceedings{li2024promptist,
  title={PROMPTIST: Automated prompt optimization for text-to-image synthesis},
  author={Li, WeiJie and Wang, Jin and Zhang, Xuejie},
  booktitle={CCF international conference on natural language processing and Chinese computing},
  pages={295--306},
  year={2024},
  organization={Springer}
}

@inproceedings{yang2023large,
  title={Large language models as optimizers},
  author={Yang, Chengrun and Wang, Xuezhi and Lu, Yifeng and Liu, Hanxiao and Le, Quoc V and Zhou, Denny and Chen, Xinyun},
  booktitle={The Twelfth International Conference on Learning Representations},
  year={2023}
}

@inproceedings{prasad2023grips,
  title={Grips: Gradient-free, edit-based instruction search for prompting large language models},
  author={Prasad, Archiki and Hase, Peter and Zhou, Xiang and Bansal, Mohit},
  booktitle={Proceedings of the 17th Conference of the European Chapter of the Association for Computational Linguistics},
  pages={3845--3864},
  year={2023}
}

@inproceedings{deng2022rlprompt,
  title={Rlprompt: Optimizing discrete text prompts with reinforcement learning},
  author={Deng, Mingkai and Wang, Jianyu and Hsieh, Cheng-Ping and Wang, Yihan and Guo, Han and Shu, Tianmin and Song, Meng and Xing, Eric and Hu, Zhiting},
  booktitle={Proceedings of the 2022 Conference on Empirical Methods in Natural Language Processing},
  pages={3369--3391},
  year={2022}
}

@inproceedings{kennedy1995particle,
  title={Particle swarm optimization},
  author={Kennedy, James and Eberhart, Russell},
  booktitle={Proceedings of ICNN'95-international conference on neural networks},
  volume={4},
  pages={1942--1948},
  year={1995},
  organization={ieee}
}

@inproceedings{lambora2019genetic,
  title={Genetic algorithm-A literature review},
  author={Lambora, Annu and Gupta, Kunal and Chopra, Kriti},
  booktitle={2019 international conference on machine learning, big data, cloud and parallel computing (COMITCon)},
  pages={380--384},
  year={2019},
  organization={IEEE}
}

@article{ghaemi2014forest,
  title={Forest optimization algorithm},
  author={Ghaemi, Manizheh and Feizi-Derakhshi, Mohammad-Reza},
  journal={Expert systems with applications},
  volume={41},
  number={15},
  pages={6676--6687},
  year={2014},
  publisher={Elsevier}
}

@article{rao2011teaching,
  title={Teaching--learning-based optimization: a novel method for constrained mechanical design optimization problems},
  author={Rao, R Venkata and Savsani, Vimal J and Vakharia, Dipakkumar P},
  journal={Computer-aided design},
  volume={43},
  number={3},
  pages={303--315},
  year={2011},
  publisher={Elsevier}
}

@article{geem2001new,
  title={A new heuristic optimization algorithm: harmony search},
  author={Geem, Zong Woo and Kim, Joong Hoon and Loganathan, Gobichettipalayam Vasudevan},
  journal={simulation},
  volume={76},
  number={2},
  pages={60--68},
  year={2001},
  publisher={Sage Publications Sage CA: Thousand Oaks, CA}
}

@article{shin2020autoprompt,
  title={Autoprompt: Eliciting knowledge from language models with automatically generated prompts},
  author={Shin, Taylor and Razeghi, Yasaman and Logan IV, Robert L and Wallace, Eric and Singh, Sameer},
  journal={arXiv preprint arXiv:2010.15980},
  year={2020}
}

@inproceedings{xu2022gps,
  title={Gps: Genetic prompt search for efficient few-shot learning},
  author={Xu, Hanwei and Chen, Yujun and Du, Yulun and Shao, Nan and Yanggang, Wang and Li, Haiyu and Yang, Zhilin},
  booktitle={Proceedings of the 2022 Conference on Empirical Methods in Natural Language Processing},
  pages={8162--8171},
  year={2022}
}

@article{singh2025agentic,
  title={Agentic retrieval-augmented generation: A survey on agentic rag},
  author={Singh, Aditi and Ehtesham, Abul and Kumar, Saket and Khoei, Tala Talaei},
  journal={arXiv preprint arXiv:2501.09136},
  year={2025}
}

@article{hong2025next,
  title={Next-generation database interfaces: A survey of llm-based text-to-sql},
  author={Hong, Zijin and Yuan, Zheng and Zhang, Qinggang and Chen, Hao and Dong, Junnan and Huang, Feiran and Huang, Xiao},
  journal={IEEE Transactions on Knowledge and Data Engineering},
  year={2025},
  publisher={IEEE}
}

@inproceedings{cheng2024black,
  title={Black-box prompt optimization: Aligning large language models without model training},
  author={Cheng, Jiale and Liu, Xiao and Zheng, Kehan and Ke, Pei and Wang, Hongning and Dong, Yuxiao and Tang, Jie and Huang, Minlie},
  booktitle={Proceedings of the 62nd Annual Meeting of the Association for Computational Linguistics (Volume 1: Long Papers)},
  pages={3201--3219},
  year={2024}
}

@inproceedings{zhang2025collaborative,
  title={Collaborative Discrete-Continuous Black-Box Prompt Learning for Language Models},
  author={Zhang, Hualin and Zhang, Haozhen and Liu, Zhekai and Gu, Bin and Chang, Yi},
  booktitle={The Thirteenth International Conference on Learning Representations},
  year={2025}
}

@article{liu2024lost,
  title={Lost in the middle: How language models use long contexts},
  author={Liu, Nelson F and Lin, Kevin and Hewitt, John and Paranjape, Ashwin and Bevilacqua, Michele and Petroni, Fabio and Liang, Percy},
  journal={Transactions of the association for computational linguistics},
  volume={12},
  pages={157--173},
  year={2024}
}

@article{sharma2025retrieval,
  title={Retrieval-Augmented Generation: A Comprehensive Survey of Architectures, Enhancements, and Robustness Frontiers},
  author={Sharma, Chaitanya},
  journal={arXiv preprint arXiv:2506.00054},
  year={2025}
}

@article{dai2022promptagator,
  title={Promptagator: Few-shot dense retrieval from 8 examples},
  author={Dai, Zhuyun and Zhao, Vincent Y and Ma, Ji and Luan, Yi and Ni, Jianmo and Lu, Jing and Bakalov, Anton and Guu, Kelvin and Hall, Keith B and Chang, Ming-Wei},
  journal={arXiv preprint arXiv:2209.11755},
  year={2022}
}

@article{wang2023query2doc,
  title={Query2doc: Query expansion with large language models},
  author={Wang, Liang and Yang, Nan and Wei, Furu},
  journal={arXiv preprint arXiv:2303.07678},
  year={2023}
}

@inproceedings{ma2023query,
  title={Query rewriting in retrieval-augmented large language models},
  author={Ma, Xinbei and Gong, Yeyun and He, Pengcheng and Zhao, Hai and Duan, Nan},
  booktitle={Proceedings of the 2023 Conference on Empirical Methods in Natural Language Processing},
  pages={5303--5315},
  year={2023}
}

@article{pan2024unifying,
  title={Unifying large language models and knowledge graphs: A roadmap},
  author={Pan, Shirui and Luo, Linhao and Wang, Yufei and Chen, Chen and Wang, Jiapu and Wu, Xindong},
  journal={IEEE Transactions on Knowledge and Data Engineering},
  volume={36},
  number={7},
  pages={3580--3599},
  year={2024},
  publisher={IEEE}
}

@article{shuster2021retrieval,
  title={Retrieval augmentation reduces hallucination in conversation},
  author={Shuster, Kurt and Poff, Spencer and Chen, Moya and Kiela, Douwe and Weston, Jason},
  journal={arXiv preprint arXiv:2104.07567},
  year={2021}
}

@article{huang2024survey,
  title={A survey on retrieval-augmented text generation for large language models},
  author={Huang, Yizheng and Huang, Jimmy},
  journal={arXiv preprint arXiv:2404.10981},
  year={2024}
}

@inproceedings{petroni2019language,
  title={Language models as knowledge bases?},
  author={Petroni, Fabio and Rockt{\"a}schel, Tim and Riedel, Sebastian and Lewis, Patrick and Bakhtin, Anton and Wu, Yuxiang and Miller, Alexander},
  booktitle={Proceedings of the 2019 conference on empirical methods in natural language processing and the 9th international joint conference on natural language processing (EMNLP-IJCNLP)},
  pages={2463--2473},
  year={2019}
}

@article{wang2022text,
  title={Text embeddings by weakly-supervised contrastive pre-training},
  author={Wang, Liang and Yang, Nan and Huang, Xiaolong and Jiao, Binxing and Yang, Linjun and Jiang, Daxin and Majumder, Rangan and Wei, Furu},
  journal={arXiv preprint arXiv:2212.03533},
  year={2022}
}

@article{geva2021did,
  title={Did aristotle use a laptop? a question answering benchmark with implicit reasoning strategies},
  author={Geva, Mor and Khashabi, Daniel and Segal, Elad and Khot, Tushar and Roth, Dan and Berant, Jonathan},
  journal={Transactions of the Association for Computational Linguistics},
  volume={9},
  pages={346--361},
  year={2021},
  publisher={MIT Press One Rogers Street, Cambridge, MA 02142-1209, USA journals-info~…}
}

@article{min2020ambigqa,
  title={AmbigQA: Answering ambiguous open-domain questions},
  author={Min, Sewon and Michael, Julian and Hajishirzi, Hannaneh and Zettlemoyer, Luke},
  journal={arXiv preprint arXiv:2004.10645},
  year={2020}
}

@inproceedings{mallen2023not,
  title={When not to trust language models: Investigating effectiveness of parametric and non-parametric memories},
  author={Mallen, Alex and Asai, Akari and Zhong, Victor and Das, Rajarshi and Khashabi, Daniel and Hajishirzi, Hannaneh},
  booktitle={Proceedings of the 61st Annual Meeting of the Association for Computational Linguistics (Volume 1: Long Papers)},
  pages={9802--9822},
  year={2023}
}

@article{jagerman2023query,
  title={Query expansion by prompting large language models},
  author={Jagerman, Rolf and Zhuang, Honglei and Qin, Zhen and Wang, Xuanhui and Bendersky, Michael},
  journal={arXiv preprint arXiv:2305.03653},
  year={2023}
}

@article{baek2025crafting,
  title={Crafting the path: Robust query rewriting for information retrieval},
  author={Baek, Ingeol and Lee, Jimin and Yang, Joonho and Lee, Hwanhee},
  journal={IEEE Access},
  year={2025},
  publisher={IEEE}
}

@article{rafailov2023direct,
  title={Direct preference optimization: Your language model is secretly a reward model},
  author={Rafailov, Rafael and Sharma, Archit and Mitchell, Eric and Manning, Christopher D and Ermon, Stefano and Finn, Chelsea},
  journal={Advances in neural information processing systems},
  volume={36},
  pages={53728--53741},
  year={2023}
}

@inproceedings{ye2025q,
  title={Q-PRM: Adaptive Query Rewriting for Retrieval-Augmented Generation via Step-level Process Supervision},
  author={Ye, Xiaopeng and Xu, Chen and Zhang, Chaoliang and Du, Zhaocheng and Xu, Jun and Wang, Gang and Dong, Zhenhua},
  booktitle={Findings of the Association for Computational Linguistics: EMNLP 2025},
  pages={15113--15128},
  year={2025}
}

@article{costa2017efficient,
  title={Efficient parameter estimation for information retrieval using black-box optimization},
  author={Costa, Alberto and Di Buccio, Emanuele and Melucci, Massimo and Nannicini, Giacomo},
  journal={IEEE Transactions on Knowledge and Data Engineering},
  volume={30},
  number={7},
  pages={1240--1253},
  year={2017},
  publisher={IEEE}
}

@article{jin2024large,
  title={Large language models on graphs: A comprehensive survey},
  author={Jin, Bowen and Liu, Gang and Han, Chi and Jiang, Meng and Ji, Heng and Han, Jiawei},
  journal={IEEE Transactions on Knowledge and Data Engineering},
  year={2024},
  publisher={IEEE}
}

\end{document}